\documentclass[twoside,11pt]{article}
\usepackage{jmlr2e}

\usepackage{thmtools} 
\usepackage{booktabs}
\usepackage{algorithm}
\usepackage{algorithmic}
\usepackage{balance}
\usepackage{doi}
\usepackage{microtype}
\usepackage{multicol}
\usepackage{enumitem}
\usepackage{tikz}
\usepackage{tikz-cd}
\usepackage{standalone}
\usepackage{tikz-3dplot}
\usepackage{textcomp}
\usetikzlibrary{patterns,arrows,arrows.meta,calc,shapes,shadows,decorations.pathmorphing,decorations.pathreplacing,automata,shapes.multipart,positioning,shapes.geometric,fit,circuits,trees,shapes.gates.logic.US,fit}
\usetikzlibrary{backgrounds,scopes}
\usepackage{balance}
\usepackage[all]{xy}
\usepackage{amsmath}
\usepackage{amssymb,mathtools}
\renewcommand{\hat}{\widehat}
\newcommand{\mM}{\mathcal{M}}
\newcommand{\pP}{\mathcal{P}}
\newcommand{\nN}{\mathcal{N}}
\newcommand{\cC}{\mathcal{C}}
\newcommand{\bO}{\mathbb{O}}
\newcommand{\bI}{\mathbb{I}}
\newcommand{\bR}{\mathbb{R}}
\newcommand{\Astar}{A}
\newcommand{\Aut}{\mathop{Aut}}
 
\usepackage[noabbrev]{cleveref}

\newcommand{\id}{\mathrm{id}}

\usepackage{xcolor}
\definecolor{yel}{RGB}{237, 164, 60}
\definecolor{mygray}{RGB}{165, 165, 165}

\ShortHeadings{Categorical semantics of compositional reinforcement learning}{Bakirtzis, Savvas, and Topcu}
\firstpageno{1}

\title{Categorical Semantics of~Compositional~Reinforcement~Learning}
\author{Georgios Bakirtzis \email
bakirtzis@telecom-paris.fr\\
\addr
LTCI, Télécom Paris, 
Institut Polytechnique de Paris
\AND
Michail Savvas
\email
michail-savvas@uiowa.edu\\
\addr
The University of Iowa
\AND
Ufuk Topcu \email
utopcu@utexas.edu\\
\addr
The University of Texas at Austin
}

\usepackage{lastpage}
\jmlrheading{26}{2025}{1-\pageref{LastPage}}{2/24; Revised
3/25}{6/25}{24-0197}{Georgios Bakirtzis, Michail Savvas, Ufuk Topcu}
\ShortHeadings{Categorical Semantics of Compositional Reinforcement Learning}{Bakirtzis, Savvas, Topcu}

\editor{Aryeh Kontorovich}
\begin{document}
\maketitle

\begin{abstract}%
Compositional knowledge representations in reinforcement learning (RL) facilitate modular, interpretable, and safe task specifications. However, generating compositional models requires the characterization of minimal assumptions for the robustness of the compositionality feature, especially in the case of functional decompositions. Using a categorical point of view, we develop a knowledge representation framework for a \emph{compositional theory} of RL. Our approach relies on the theoretical study of the category $\mathsf{MDP}$, whose objects are Markov decision processes (MDPs) acting as models of tasks. The categorical semantics models the compositionality of tasks through the application of \emph{pushout} operations akin to combining puzzle pieces. As a practical application of these pushout operations, we introduce \emph{zig-zag} diagrams that rely on the compositional guarantees engendered by the category $\mathsf{MDP}$. We further prove that properties of the category $\mathsf{MDP}$ unify concepts, such as enforcing safety requirements and exploiting symmetries, generalizing previous abstraction theories for RL. %

\end{abstract}%
\begin{keywords}
universal properties, Markov decision processes, category theory  
\end{keywords}

\section{Introduction}

Verification of reinforcement learning (RL) systems is essential for mastering increasingly complex tasks, ensuring reliable and trustworthy behavior, and mitigating the risk of hazardous control actions in dynamic environments~\citep{jothimurugan:2021,gur:2021,li:2021,szabo:2012}. The construction of compositional theories for Markov decision processes (MDPs) verifies the overall behavior of an RL system by parts, enabling the segmentation into manageable components. Exploiting the compositionality feature in MDPs suggests that regardless of the data or flexibility of the  RL algorithms we deploy, it is beneficial to model using modular task specifications and precisely define the structural features of sequential decision-making problems such that the engineered artifact is interpretable.

In particular, functional compositionality is essential to engendering increasingly adaptable systems by providing an interpretable language for specifying tasks. However, methods that functionally compose problem formulations are scarce~\citep{mendez:2022}. Functional compositionality \emph{predicts} a system's behavior by the combination of behaviors deriving from its parts and the combination \emph{preserves} properties emerging from the parts~\citep{genovese:2018}. To extend this line of research, we construct a unifying compositional theory for engineering RL systems that mechanizes functional composition into subprocess behaviors.  

\begin{figure}[!t]
    \centering
    \begin{tikzpicture}
  \clip (-4.25cm, -1.6cm) rectangle (2.85cm, 2.4cm);

  \node[rotate=-90, minimum size=.5cm, regular polygon, regular polygon sides=3, draw=yel, fill=yel, scale=0.5] (triangle) {};

  \node[text width=2.5cm, align=right] (decomposability) at ([xshift=-3cm]triangle.center) {Decomposability};
  \node[text width=2.5cm, align=left] (usage) at ([xshift=2.5cm]triangle.center) {Usage};
  
  \draw[gray] ([yshift=0.1cm,xshift=0.2cm]decomposability.north) -- ([yshift=0.6cm,xshift=0.2cm]decomposability.north) node [midway, above, yshift=0.5cm, align=center,text width=2cm, font=\footnotesize] {factored \\ relational \\ modular};
  \draw[gray] ([xshift=-0.7cm,yshift=0.1cm]usage.north) -- ([xshift=-0.7cm,yshift=0.6cm]usage.north) node [midway, above, yshift=0.5cm, align=center,text width=2cm, font=\footnotesize] {interpretable\\ safe \\ general};
  
  \node at ([yshift=-1cm, xshift=0.1cm]triangle.south) {\textcolor{yel}{Structure}};
\end{tikzpicture}
    \caption{Structure improves interpretability, safety, and generalization (adapted from \citet{mohan:2023}).}
    \label{fig:structure}
\end{figure}

To achieve this explicit definition of compositionality in RL, we give a symbolic and semantic interpretation of compositional phenomena of the problem or task by translating them into categorical properties. Category theory is well-suited for posing and solving \emph{structural} problems. As such, categorical constructions, like \emph{pullbacks} and \emph{pushouts}, describe general operations that model (de)compositions of objects, thereby giving formal meaning to clusters of objects as a composite and how they interface with each other. This formalization is not merely a theoretical exercise; it can improve RL systems' interpretability, safety, and generalization capabilities. Interpretability is augmented as these categorical structures provide a systematic framework for understanding the relationships and interactions within an RL system. Safety is addressed through precisely controlling and predicting these interactions, especially in complex or composite systems where emergent behaviors might pose risks. Thus, through a compositional theory it is possible to enhance the core aspects of RL systems---making them interpretable, safer, and adaptable to diverse applications (Figure~\ref{fig:structure}).

The notion of MDP is the structural instantiation of a general sequential decision-making problem~\citep{ok:2018}. We study compositionality from the perspective of subprocess MDPs and examine compositional properties governing MDPs that result in universal interfaces. In particular, addressing task (de)composition requires operationalizing~\citep {todorov:2009} a given overall behavior. In this study, categorical semantics makes the types of MDP operations in different configurations explicit, extending the algebraic approach~\citep{perny:2005}. While hierarchical MDPs also allow for operationalizing tasks in RL~\citep{parr:1997,dietterich:2000,nachum:2018,zheng:2020,klissarov:2025}, the categorical semantics defines the exorable and incorporeal forms of abstraction, refinement, and compositional operations through universal properties. Universal properties are particularly desirable for knowledge representations because they govern the definition of new admissible composite systems, thereby constructively producing modular components. In the compositional theory, categories define a precise \emph{syntax} for MDP instantiations and assign \emph{semantics} to particular compositional behaviors (Section~\ref{sec:algebra}).

We construct a categorical semantics for sequential task completion problems. The first superposes obstacles in a grid world using the operation of pullbacks (Section~\ref{app:grid}). The second exploits the (potential) symmetric structure of RL problems using the operation of a pushout (Section~\ref{sec:sym}). The third is a design of compositional task completion, where zig-zag diagrams synthesize the behavior of a fetch-and-place robot (Section~\ref{sec:design}). Zig-zag diagrams (Section~\ref{sub:zig-zag}) denote composite MDPs produced by gluing together subprocess MDPs, preserving the relationship between types of actions and states. While this paper aims to develop a \emph{compositional theory} of RL, our constructions have a computational interpretation that partitions problem spaces and manages the increasing complexity of RL systems~\citep{bakirtzis:2024c}.\\

\noindent \emph{Conventions} \, Subscripts for MDP elements refer to a particular instantiation of the definition. For example, $S_\nN$ refers to the state space of the MDP $\nN$. Certain preliminary concepts are introduced in Appendix~\ref{app:cat}. Unless otherwise noted, we consider probability measures on spaces expressed by a probability density function with respect to a given ambient probability measure when clear from context.

\section{A Compositional Theory of Reinforcement Learning}
\label{sec:algebra}

In this section, we introduce the unit MDP and the latent structure
of the environments in this work. We then introduce various compositionality features of MDP structures through a theoretical study of categorical properties.

\subsection{The Category \texorpdfstring{$\mathsf{MDP}$}{\textsf{MDP}}}

\begin{definition}[MDP]\label{def:mdp}
An MDP $\mM =(S, \Astar, \psi, T, R)$ is a $5$-tuple consisting of the following data:
\begin{itemize}
    \item A measurable space $S$ with a fixed (implicit) $\sigma$-algebra, called the \textit{state space} of $\mM$.
    \item A set $A$, called the state-action space of $\mM$. This is the set of actions available at all different states of $S$.%
    \item A %
    function $\psi \colon A \to S$ that maps an action $a \in \Astar$ to its associated state $s \in S$. 
    \item A function $T \colon \Astar \to \pP_S$, where $\pP_S$ denotes the space of probability measures on $S$. This is the information of the transition probabilities on $\mM$.
    \item A reward function $R \colon A \to \bR$.
\end{itemize}
\end{definition}

The actions an agent can take at a particular state $s$ are given by the set $\psi^{-1}(s) \subseteq \Astar$. We denote this set of available actions for each state $s$ by $\Astar_s$. Knowing the action spaces $\Astar_s$ for all $s \in S$, one may recover $\Astar$ and $\psi$ by the assignment
\begin{align} \label{action space from fibers}
    \Astar = \coprod_{s \in S} \Astar_s,\ \psi \colon \Astar_s \mapsto s \in S.
\end{align}

If the actions of the MDP are not state-specific and uniform across the whole state space $S$, then we can take $\Astar = S \times A_0$ for a fixed set of actions $A_0$ and $\psi \colon \Astar \to S$ is then just the projection onto the first factor. 

    One departure of the above definition from standard definitions in the literature is the added flexibility in allowing the actions to vary across the state space $S$. While many MDPs are uniform and thus the state-action space is naturally a product $A = S \times A_0$ as in the previous remark, there are natural MDP environments where this uniformity fails. For example, an agent moving inside a maze will not be able to move in all four directions when faced with a wall or another obstacle, forming a boundary which can vary within the environment defining the state space.

    Another possible simplification occurs when the MDP rewards only depend on the state and not the action taken. In that case, the reward function $R$ factors through $\psi$ and a state-dependent reward function $R_0 \colon S \to \bR$, so that $R = R_0 \circ \psi$.

Having defined MDPs, we now proceed to define morphisms to obtain the corresponding category whose objects are MDPs (Definition~\ref{def:mdp}). 

\begin{definition}[Category of MDPs]
MDPs form a category $\mathsf{MDP}$ whose morphisms are defined as follows: 

Let $\mM_i = (S_i, \Astar_i, \psi_i, T_i, R_i)$, with $i=1,2$, be two MDPs. 

A morphism $m = (f,g) \colon \mM_1 \to \mM_2$ is the data of a measurable function $f \colon S_1 \to S_2$ and a function $g \colon \Astar_1 \to \Astar_2$ satisfying the following compatibility conditions: 

\begin{enumerate}
    \item The diagram
    \begin{align} \label{action-space compat diagram}
        \xymatrix{
        \Astar_1 \ar[r]^-{g} \ar[d]_-{\psi_1} & \Astar_2 \ar[d]^-{\psi_2} \\
        S_1 \ar[r]_-{f} & S_2}
    \end{align}
    is commutative.
\end{enumerate}

\begin{enumerate}
    \item[2.] The diagram 
    \begin{align} \label{action-prob compat diagram}
        \xymatrix{
        \Astar_1 \ar[r]^-{g} \ar[d]_-{T_1} & \Astar_2 \ar[d]^-{T_2} \\
        \pP_{S_1} \ar[r]_-{f_*} & \pP_{S_2}}
    \end{align}
    is commutative, where $f_*$ maps a probability measure $\mu_1 \in \pP_{S_1}$ to the pushforward measure $\mu_2 = f_* \mu_1 \in \pP_{S_2}$ under $f$.
\end{enumerate}

\begin{enumerate}
    \item[3.] The diagram 
    \begin{align} \label{reward compat diagram}
        \xymatrix{
        \Astar_1 \ar[r]^-{g} \ar[dr]_-{R_1} & \Astar_2 \ar[d]^-{R_2} \\
         & \bR}
    \end{align}
    is commutative.
\end{enumerate}
\end{definition}

The commutative diagrams above express the compatibility between two MDPs in the sense that their interfaces and rewards agree. 

Namely, diagram~\eqref{action-space compat diagram} guarantees that if an action $a_1$ in MDP $\mM_1$ is associated to a state $s_1 \in S_1$, then its image action $a_2=g(a_1)$ under the morphism $m$ is associated to the image state $s_2=f(s_1)$. 

Similarly, diagram~\eqref{action-prob compat diagram} ensures that the transition probability from any state $s_1$ to the measurable subset $f^{-1}(U) \subseteq S_1$ after taking action $a_1$ in $\mM_1$, where $U \subseteq S_2$ is any measurable subset, is equal to the transition probability from the state $s_2=f(s_1)$ to $U \subseteq S_2$ under action $a_2 = g(a_1)$ in $\mM_2$.

Finally, diagram~\eqref{reward compat diagram} ensures that the reward obtained after taking action $a_1$ at the state $s_1$ in MDP $\mM_1$ equals the reward obtained after taking action $a_2 = g(a_1)$ at the state $s_2 = f(s_1)$ in MDP $\mM_2$.

Some immediate features of the category $\mathsf{MDP}$ are as follows.

\begin{definition}[Empty and constant MDP] \label{def:pt}
    The empty MDP $\emptyset$ is the MDP whose state space and action spaces are the empty set. 
    The constant MDP $\mathsf{pt}$ is the MDP whose state space and action spaces are the one-point set. 
\end{definition}

\begin{remark}
    The constant MDP is not unique, as the reward function value can be defined to be equal to any number. We refer to any such instance as the constant MDP.
\end{remark}

The category $\mathsf{MDP}$ has an initial object and a collection of partially terminal objects.

\begin{proposition} \label{prop:empty-const}
    For every MDP $\mM$, there exists a unique morphism $\emptyset \to \mM$. In particular, $\emptyset$ is the initial object of $\mathsf{MDP}$.
    Every MDP $\mM$ with a constant reward function admits a unique, natural morphism $\mM \to \mathsf{pt}$. %
\end{proposition}

\noindent
\emph{Subprocesses} \,
There is also a natural definition of subobjects in $\mathsf{MDP}$. In particular, the definition of morphisms can correctly capture the notion of a subprocess of an MDP. 
\begin{definition}[Subprocesses]\label{subprocess}
We say that the MDP $\mM_1$ is a subprocess of the MDP $\mM_2$ if there exists a morphism $m = (f,g) \colon \mM_1 \to \mM_2$ where $f$ and $g$ are injective, that is the morphism $m$ is injective on state and state-action spaces.

We say that a subprocess $\mM_1$ is a full subprocess of $\mM_2$ if diagram~\eqref{action-space compat diagram} is cartesian. Equivalently, this means that $A_1 = \psi_2^{-1}(f(S_1))$ as a subset of $A_2$. %
\end{definition}

Since $f$ is injective, we may consider the state space $S_1$ as a subset of $S_2$. Moreover, the condition that diagram~\eqref{action-space compat diagram} is cartesian means that $S_1$ is closed under actions in $\mM_2$ and the state-action spaces of $\mM_1$ and $\mM_2$ coincide over $S_1$. Thus, $\mM_1$ being a full subprocess of $\mM_2$ implies that an agent following the MDP $\mM_2$ who finds themself at a state $s_1 \in S_1$ will remain within $S_1$ no matter which action $a_1 \in \Astar_1$ they elect to take.

Conversely, for a MDP $\mM_2$ and any subset $S_1 \subseteq S_2$ there is a canonical subprocess $\mM_1$ with state space $S_1$, whose action space $\Astar_1$ is defined as
\begin{align} \label{def of A_1}
    \Astar_1 \coloneqq \psi_2^{-1}(S_1) \cap T_2^{-1}( f_\star (\pP_{S_1})).
\end{align}

In fact, $\mM_1$ is uniquely characterized as the maximal such subprocess.

\begin{proposition}
Any subprocess $\mM_1' \to \mM_2$ with state space $S_1$ factors uniquely through the subprocess $\mM_1 \to \mM_2$.
\end{proposition}

\begin{proof}
This follows from the fact that any action $a_2 \in (A_2)_{s_1}$ such that $T_2(a_2)$ is a measure supported on $S_1$ gives an element of $A_1$, as defined in equation~\eqref{def of A_1}.
\end{proof}

In the ensuing subsections, we will investigate further, more complicated features of the category $\mathsf{MDP}$. In particular, we will be interested in appropriate and universal constructions of products and ``unions" of MDPs.

\subsection{Pullbacks: Universal Interfaces}

Interesting categorical properties are \emph{universal}. Universal properties represent specific ideals of behavior within a defined category~\citep{spivak:2014,asperti:1991}. A pullback (or fiber product) is a categorical generalization of the notion of cartesian product whose universal property is determined by it being maximal in a certain sense. Besides the cartesian product, another common example is the intersection $S_1 \cap S_2$ of two subsets $S_1, S_2 \subseteq S_3$, obtained as the pullback of the two inclusions of sets $S_1 \to S_3$ and $S_2 \to S_3$.

We prove that well-formed MDPs have partial pullbacks in the category $\mathsf{MDP}$, meaning that there exists a way to compose the data of one MDP with another to create an interface between MDPs.

The intuition behind a pullback of MDPs is the idea of \emph{intersecting} two MDPs $\mM_1$ and $\mM_2$ viewed as components of a third MDP $\mM_3$ through morphisms $m_1 \colon \mM_1 \to \mM_3$ and $m_2 \colon \mM_2 \to \mM_3$. In general, the morphisms $m_i$ do not need to define subprocesses and this allows for valuable flexibility, ranging from intersections to products. For example, the cartesian product $\mM_1 \times \mM_2$ of two MDPs will be obtained when $\mM_3 = \mathsf{pt}$ is the constant MDP (Definition~\ref{def:pt}). In the other extreme, we can obtain intersections of subprocesses, as for the grid world problem, which will be studied later on in this paper.

Write $\mM_i = (S_i, A_i, \psi_i, T_i, R_i)$ for $i=1,2,3$. The structure of the state space of the pullback as a measure space requires some care in the construction and is the reason for the failure of the existence of a fully universal pullback. If this were indeed possible, one would then be able to deduce a construction of pullbacks for the category of measurable spaces, which is known not to exist.

However, we will succeed, allowing for the weakening of the universal properties. We will obtain the following theorem by examining progressively more general cases. %

\begin{theorem}
Let $\mM_i = (S_i, A_i, \psi_i, T_i, R_i)$ for $i=1,2,3$ be MDPs. There exists an MDP $\mM = \mM_1 \times_{\mM_3} \mM_2$ with state space $S = S_1 \times_{S_3} S_2$ (defined in equation~\ref{eq:state space of fiber product}) and state-action space $A = A_1 \times_{A_3} A_2$ (defined in equation~\ref{eq:action space of fiber product}) which fits in a commutative diagram in $\mathsf{MDP}$:
\begin{align*}
    \xymatrix{
    \mM \ar[d] \ar[r] & \mM_1 \ar[d]^-{m_1} \\
    \mM_2 \ar[r]_-{m_2} & \mM_3 
    }
\end{align*}
The MDP $\mM$ is universal with respect to diagrams whose morphisms are conditionally independent (Definition~\ref{def:cond indep}).
\end{theorem}

We establish the theorem in steps, which we also discuss as we go. 

Suppose that we have three MDPs $\mM_1, \mM_2, \mM_3$ together with two morphisms $m_1 = (f_1, g_1) \colon \mM_1 \to \mM_3$ and $m_2 = (f_2, g_2) \colon \mM_2 \to \mM_3$.

We begin by defining the pullback's state, action spaces, and rewards and continue with the setup of the transition probabilities.

\smallskip
\noindent
\emph{Construction of states, actions and, rewards} \,
For the state and action spaces, we set
\begin{align}
    S & = S_1 \times_{S_3} S_2 = \lbrace (s_1, s_2) \in S_1 \times S_2 \ \vert \ f_1(s_1) = f_2(s_2) \in S_3 \rbrace, \label{eq:state space of fiber product}\\ 
    \Astar & = \Astar_1 \times_{\Astar_3} \Astar_2= \lbrace (a_1, a_2) \in A_1 \times A_2 \ \vert \ g_1(a_1) = g_2(a_2) \in A_3 \rbrace. \label{eq:action space of fiber product} 
\end{align}
By standard properties of pullbacks (of sets), the maps $\psi_i \colon \Astar_i \to S_i$ for $i=1,2,3$ induce a canonical morphism $\psi \colon \Astar \to S$. Write $pr_i \colon S \to S_i$ and $\rho_i \colon \Astar \to \Astar_i$ for the projection maps, where $i=1,2$.

Since we need $S$ to be a measurable space, we endow it with the $\sigma$-algebra generated by all subsets 
$$pr_1^{-1}(U_1) \cap pr_2^{-1}(U_2),$$
where $U_1, U_2$ are any two measurable subsets of $S_1$ and $S_2$ respectively. The projections $pr_i$ are then tautologically measurable functions and this is the minimal choice of $\sigma$-algebra for which this holds.

For the reward function, we set $R = R_3 \circ g_1 \circ pr_1 = R_3 \circ g_2 \circ pr_2 \colon A \to \bR$.

This minimal $\sigma$-algebra on $S$ is potentially small. However, when $\sigma$-algebras of $S_1$ and $S_2$ are their power sets (for example, in the case where $S_1$ and $S_2$ are finite or discrete measure spaces), then the $\sigma$-algebra is also the power set of $S$, so we do not view this potential smallness as an issue. 

\smallskip
\noindent
\emph{Transition probabilities} \, Let now $a \in \Astar$ with projections $a_1 \in \Astar_1$ and $a_2 \in \Astar_2$ mapping in turn to an action $a_3 \in \Astar_3$. For brevity, write $\mu_i = T_i(a_i) \in \pP_{S_i}$ for $i=1,2$ and $\mu_3 = (f_i)_\star \mu_i \in \pP_{S_3}$. 

Our goal is to construct $\mu = T(a) \in \pP_S$ and obtain for each index $i=1,2$ a commutative diagram
\begin{align*}
    \xymatrix{
    \Astar \ar[d]_-{T} \ar[r]^-{\rho_i} & \Astar_i \ar[d]^-{T_i} \\
    \pP_S \ar[r]_-{(pr_i)_\star} & \pP_{S_i}.
    }
\end{align*}
Having achieved that, we will get, by definition, an MDP $\mM$ fitting in the commutative diagram
\begin{align} \label{fiber product of MDP}
    \xymatrix{\mM \ar[r]^-{(pr_1, \rho_1)} \ar[d]_-{(pr_2, \rho_2)} & \mM_1 \ar[d]^-{m_1} \\
    \mM_2 \ar[r]_-{m_2} & \mM_3.}
\end{align}
We would moreover like this diagram to be universal among commutative diagrams of the form
\begin{align} \label{commutative square of MDP}
    \xymatrix{\nN \ar[r]^-{(\alpha_1, \beta_1)} \ar[d]_-{(\alpha_2, \beta_2)} & \mM_1 \ar[d]^-{m_1} \\
    \mM_2 \ar[r]_-{m_2} & \mM_3,}
\end{align}
in the sense that any commutative diagram~\eqref{commutative square of MDP} should be induced by a canonical morphism $\nN \to \mM$ and composing with the projection maps $\mM \to \mM_1$ and $\mM \to \mM_2$. This will not generally be possible and we will need to restrict our attention to pairs of morphisms $\mM_1 \to \mM_3,\ \mM_2 \to \mM_3$ satisfying a certain independence condition, as we will see below. For such pairs of morphisms, we can indeed achieve our goal.

\medskip
We ask: \vspace{-.5em}
\begin{enumerate}
    \item[\textlangle Q1\textrangle] Under what conditions does $\mu$ exist?
    \item[\textlangle Q2\textrangle] If $\mu = T(a)$ exists for all $a \in A$, what kind of universality properties does $\mM$ have? 
\end{enumerate}

We describe several instances in which the measure $\mu = T(a)$ can be constructed, in order of increasing generality. We begin with the case where one of the two morphisms is a subprocess, then move on to cases with discrete state spaces, and finally conclude with the most general case of local fibrations. In addition, we obtain a simple-to-state and clean universality statement in the subprocess and discrete cases using conditional independence.

\smallskip
\noindent
\emph{The subprocess case} \,
Suppose that one of $\mM_1 \to \mM_3$ and $\mM_2 \to \mM_3$ is a subprocess. Assume this is the case for $\mM_2 \to \mM_3$ without loss of generality.
By Definition~\ref{subprocess}, this implies that the maps $pr_1 \colon S \to S_1$ and $\rho_1 \colon \Astar \to \Astar_1$ are injective. 
Moreover, since $(f_1)_\star \mu_1 = (f_2)_\star \mu_2 = \mu_3$, it follows that $\mu_1$ is supported on $S$. We define $T(a) = \mu_1$ as a measure on $S$.

\begin{proposition}
The diagram~\eqref{fiber product of MDP} is universal among diagrams of the form~\eqref{commutative square of MDP} for which the morphism $\nN \to \mM_1$ defines a subprocess.
\end{proposition}

\begin{proof}
Suppose that $(\alpha_1, \beta_1) \colon \nN \to \mM_1$ defines a subprocess. We have induced morphisms $\alpha=(\alpha_1, \alpha_2) \colon S_\nN \to S=S_1 \times_{S_3} S_2$ and $\beta=(\beta_1, \beta_2) \colon A_\nN \to A = A_1 \times_{A_3} A_2$.

Both maps are injective since the compositions $pr_1 \circ \alpha = \alpha_1$ and $\rho_1 \circ \beta = \beta_1$ are injective.

For any $a \in A_\nN$ we need to check that $\alpha_\star T_\nN (a) = T(\beta(a))$.

But $T_1(\beta_1(a))$ is supported on $\nN$ and we have $T_1(\beta_1(a)) = T(\beta(a))$. 

On the other hand, $(pr_1)_\star \alpha_\star T_\nN (a) = (pr_1 \circ \alpha)_\star T_\nN (a) = (\alpha_1)_\star T_\nN (a) = T_1 (\beta_1(a))$, as wanted.
\end{proof}

\noindent
\emph{The discrete case} \,
Assume that the spaces $S_1, S_2, S_3$ are discrete or finite and their $\sigma$-algebras are their power sets.

Consider the function $\nu \colon S_1 \times_{S_3} S_2 \to \mathbb{R}$ defined by the formula
\begin{equation} \label{definition of nu}
    \nu(s_1, s_2) = \begin{cases} \frac{\mu_1(s_1) \mu_2(s_2)}{\mu_3(f_1(s_1))} = \frac{\mu_1(s_1) \mu_2(s_2)}{\mu_3(f_2(s_2))},  &\text{ if } \mu_3(f_1(s_1)) = \mu_3(f_2(s_2)) > 0, \\  0, & \text{ if } \mu_3(f_1(s_1))= \mu_3(f_2(s_2)) = 0. \end{cases}
\end{equation}

We define the measure $\mu = T(a) \in \pP_S$ to be the one with probability density function $\nu$. Thus,
\begin{align} \label{product measure for N}
   \mu \left( pr_1^{-1}(U_1) \cap pr_2^{-1}(U_2) \right) \coloneqq \sum_{(s_1, s_2) \in  pr_1^{-1}(U_1) \cap pr_2^{-1}(U_2)} \mathclap{\nu(s_1, s_2).} %
\end{align}

This satisfies the desired properties to define MDP morphisms $\mM \to \mM_i$, $i=1,2$ (see Proposition~\ref{pushforward works} below). We will now show that it is also universal in an appropriate sense.

\begin{definition}\label{def:cond indep}
    We say that two morphisms $(\alpha_i, \beta_i) \colon \nN \to \mM_i$, $i=1,2$, in $\mathsf{MDP}$ form a \emph{conditionally independent pair with respect to $\mM_3$} if for any $a \in \Astar_\nN$ with images $a_i = \beta_i (a) \in \Astar_i$ and any measurable subsets $U_i \subseteq S_i$, we have
\begin{align} \label{condition for independence}
    T_\nN(a) \left( \alpha_1^{-1}(U_1) \cap \alpha_2^{-1}(U_2) \right) = \sum_{(s_1,s_2) \in pr_1^{-1}(U_1) \cap pr_2^{-1}(U_2)} \mathclap{\nu(s_1,s_2).} 
\end{align}
\end{definition}
This assignment determines the measure $T_\nN$ only on the $\sigma$-algebra generated by sets $\alpha_1^{-1}(U_1) \cap \alpha_2^{-1}(U_2)$.

Our definition of conditional independence is consistent with the corresponding notion considered by~\citet{swart:1996}.
We can now prove a universality statement.

\begin{proposition}\label{prop:discrete fiber prod thm}
The morphisms $\mM \to \mM_1$ and $\mM \to \mM_2$ in diagram~\eqref{fiber product of MDP} form a conditionally independent pair. Moreover, diagram \eqref{fiber product of MDP} is universal among diagrams~\eqref{commutative square of MDP} where the morphisms $\nN \to \mM_i$ form a conditionally independent pair.
\end{proposition}

\begin{proof}
The independence of $\mM \to \mM_1$ and $\mM \to \mM_2$ follows immediately by observing that equations~\eqref{product measure for N} and \eqref{condition for independence} coincide.

For the universality statement, suppose as above that we have two independent morphisms $(\alpha_i, \beta_i) \colon \nN \to \mM_i$, $i=1,2$, in $\mathsf{MDP}$, that fit in a commutative square~\eqref{commutative square of MDP}.

Since $\mM$ has state and action spaces given by $S=S_1 \times_{S_3} S_2$ and $\Astar = \Astar_1 \times_{\Astar_3} \Astar_2$ respectively and the diagram gives that $f_1 \circ \alpha_1 = f_2 \circ \alpha_2$ and $g_1 \circ \beta_1 = g_2 \circ \beta_2$, we get canonical morphisms $\gamma \colon S_{\nN} \to S$ and $\delta \colon \Astar_{\nN} \to \Astar$. These satisfy
$$ pr_i \circ \gamma = \alpha_i, \quad \rho_i \circ \delta = \beta_i,\quad i=1,2,$$
and moreover by construction $\gamma \circ \psi_{\nN} = \psi \circ \delta$.

To obtain a morphism $\nN \to \mM$, it remains to check that the corresponding diagram~\eqref{action-prob compat diagram} commutes, as compatibility of rewards is clear. For $a \in \Astar_\nN$, write $\beta_i(a) = \rho_i (\delta(a)) = a_i \in \Astar_i$. We then have, using formula~\eqref{condition for independence} for the third equality and formula~\eqref{product measure for N} for the last equality,
\begin{align*}
    &\gamma_\star T_{\nN}(a) \left( pr_1^{-1}(U_1) \cap pr_2^{-1}(U_2) \right) = \\ &= T_{\nN}(a) \left( \gamma^{-1}(pr_1^{-1}(U_1)) \cap \gamma^{-1}(pr_2^{-1}(U_2)) \right) =\\
    & = T_{\nN}(a) \left( \alpha_1^{-1}(U_1) \cap \alpha_2^{-1}(U_2) \right) =\\
    & = \sum_{(s_1,s_2) \in pr_1^{-1}(U_1) \cap pr_2^{-1}(U_2)} \mathclap{\nu(s_1,s_2) =}\\
    & = T(\delta(a)) \left( pr_1^{-1}(U_1) \cap pr_2^{-1}(U_2) \right),
\end{align*}
which implies that $\gamma_\star \circ T_\nN = T \circ \delta$, completing the proof.
\end{proof}

\noindent
\emph{The case of $S_3$ finite} \,
Suppose that $S_3$ is finite with weights for the measure $\mu$ given by a density function $\nu_3$.

Suppose in addition that $S_1, S_2$ are measurable subsets of ambient measure spaces $(K_1, \tau_1), (K_2, \tau_2)$ with density functions $\nu_1, \nu_2$ so that for any measurable subset $U_i \subseteq S_i$ we have $\mu_i(U_i) = \int_{U_i} \nu_i \ \mathrm{d} \tau_i$. This is not restrictive and we allow it for exhibition and consistency with the preceding case. One can always take $\nu_i$ to be identically $1$ and then $\mu_i = \tau_i$.

As above, we consider the function  $\nu \colon S_1 \times_{S_3} S_2 \to \mathbb{R}$,
\[
   \nu(s_1, s_2) = \begin{cases} 
   \frac{\nu_1(s_1) \nu_2(s_2)}{\nu_3(f_1(s_1))} = \frac{\nu_1(s_1) \nu_2(s_2)}{\nu_3(f_2(s_2))}, & \text{ if } \nu_3(f_1(s_1)) = \nu_3(f_2(s_2)) > 0 \\  0,  & \text{ if } \nu_3(f_1(s_1))= \nu_3(f_2(s_2))=0, 
   \end{cases}
\]
and define the measure $\mu=T(a) \in \pP_S$ to be the one with probability density $\nu$ with respect to the product measure $\tau_1 \otimes \tau_2$ on $K_1 \times K_2$ so that
\begin{align} \label{product measure}
   \mu \left( pr_1^{-1}(U_1) \cap pr_2^{-1}(U_2) \right) \coloneqq \int_{pr_1^{-1}(U_1) \cap pr_2^{-1}(U_2)} \nu \ \mathrm{d} (\tau_1 \otimes \tau_2). %
\end{align}

 The case of $S_3$ finite encompasses the discrete case above, since for $S_1, S_2$ we may take the ambient space $K$ to be $S_1 \coprod S_2$ with measure given by set cardinality. We treat the discrete case separately because we can obtain a clean and simpler universality statement, which we hope clarifies the exposition.

\begin{proposition} \label{pushforward works}
For the probability measure $\mu$, we have $(pr_i)_\star \mu = \mu_i$ for $i=1,2$.
\end{proposition}

\begin{proof}
We first observe that for any $s_3 \in S_3$
\begin{align*}
    \nu_3 (s_3) & = \mu_3 (s_3) \\ 
    & =  (f_2)_\star \mu_2 (s_3) \\
    & = \mu_2(f_2^{-1}(s_3)) \\
    & = \int_{f_2^{-1}(s_3)} \nu_2 (s_2) \ \mathrm{d} \tau_2.
\end{align*}

We then have for any measurable subset $U_1 \subseteq S_1$
\begin{align*}
    (pr_1)_\star \mu(U_1) & = \mu(pr_1^{-1}(U_1)) = \mu(U_1 \times_{S_3} S_2) \\
    & = \int_{U_1 \times_{S_3} S_2} \nu \ \mathrm{d} (\tau_1 \otimes \tau_2)\\ 
    & = \int_{U_1 \times S_2} \bI_{U_1 \times_{S_3} S_2} \cdot \nu \ \mathrm{d} (\tau_1 \otimes \tau_2),
\end{align*}
so, applying Fubini--Tonelli's theorem, we obtain
\begin{align*}
    \mu(U_1 \times_{S_3} S_2) & = \int_{U_1} \left( 
    \int_{S_2} \bI_{U_1 \times_{S_3} S_2} \cdot \nu \ \mathrm{d} \tau_2 \right) \mathrm{d} \tau_1 \\
    & = \int_{U_1} \nu_1 (s_1) \left( 
    \int_{f_2^{-1}(f_1(s_1))} \frac{\nu_2(s_2)}{\nu_3(f_2(s_2))} \ \mathrm{d} \tau_2 \right) \mathrm{d} \tau_1 \\
    & = \int_{U_1} \nu_1 (s_1) \left( 
    \int_{f_2^{-1}(f_1(s_1))} \frac{\nu_2(s_2)}{\nu_3(f_1(s_1))} \ \mathrm{d} \tau_2 \right) \mathrm{d} \tau_1 \\
    & = \int_{U_1} \frac{\nu_1(s_1)}{\nu_3(f_1(s_1))} \left( 
    \int_{f_2^{-1}(f_1(s_1))} \nu_2(s_2) \ \mathrm{d} \tau_2 \right) \mathrm{d} \tau_1 \\
    & = \int_{U_1} \frac{\nu_1(s_1)}{\nu_3(f_1(s_1))} \nu_3(f_1(s_1)) \ \mathrm{d} \tau_1 \\
    & = \int_{U_1} {\nu_1(s_1)} \ \mathrm{d} \tau_1 = \mu_1(U_1).
\end{align*}
This shows that $(pr_1)_\star \mu = \mu_1$. An identical computation shows that $(pr_2)_\star \mu = \mu_2$ and we are done.
\end{proof}

\begin{remark}\label{rem:pt}
Suppose that we consider MDPs without a reward function. In that case, when $\mM_3$ is the terminal object $\mathsf{pt}$ and we take $S_1 = K_1$, $S_2 = K_2$ and $\nu_1$, $\nu_2$ identically $1$, we obtain the cartesian product $\mM_1 \times \mM_2$ of two MDPs $\mM_1$ and $\mM_2$. One may then equip this product with a non-unique choice of reward function, possibly depending on context. A common such choice would be the sum of the reward functions of $\mM_1$ and $\mM_2$, but any other function of the two reward functions would work as well.

In the presence of rewards, the situation is more subtle, as obtaining the above cartesian product through a pullback over $\mathsf{pt}$ requires that the reward functions of $\mM_1$ and $\mM_2$ are constant and have the same value.
\end{remark}

\smallskip
\noindent
\emph{The local fibration case} \,
We finally treat the general case. We introduce some terminology.

\begin{definition}
A measurable function between two measurable spaces $f \colon X \to Y$  is a local fibration if there exists a measurable partition $Y = Y_1 \coprod \cdots \coprod Y_N$ such that for every index $i$, we have that $X_i = f^{-1}(Y_i) \simeq F_i \times Y_i$ for some measure space $F_i$ and $f|_{X_i}$ is projection onto $Y_i$.

A morphism $(f,g) \colon \mM_1 \to \mM_2$ between MDPs is called a local fibration if the map $f \colon S_1 \to S_2$ is a local fibration.
\end{definition}

The intuition behind the introduction of a local fibration is that it allows us to practically treat the base state space $S_3$ as finite, essentially reducing the case of a continuous state space to that of a discrete state space.

We now generalize Definition~\ref{def:cond indep} to the local fibration setting, following the definition of conditional independence given by~\cite{swart:1996}.

\begin{definition}\label{condind}
Let $(\alpha_i, \beta_i) \colon \nN \to \mM_i$, $i=1,2$, be two morphisms fitting into a commutative diagram
\begin{align*}
    \xymatrix{
    \nN \ar[r] \ar[d] & \mM_1 \ar[d] \\
    \mM_2 \ar[r] & \mM_3,}
\end{align*}
where the two morphisms $\mM_1 \to \mM_3$ and $\mM_2 \to \mM_3$ are local fibrations.

For any $a \in A_\nN$, write $a_i = \beta_i (a) \in A_i$ and $\mu_i = T_i(a_i) \in \pP_{S_i}$ for $i=1,2$. Let also $\mu_3 = (f_i)_\star \mu_i \in \pP_{S_3}$ for $i=1,2$. Then, by possibly refining partitions, we can assume that $S_3 = Z_1 \coprod \cdots \coprod Z_N$ is a common partition for the fibration structure of the maps $f_i$ with respect to the measures determined by $a \in \Astar_\nN$, so that $S_1 = X_1 \times Z_1 \coprod \cdots \coprod X_N \times Z_N$, $S_2 = Y_1 \times Z_1 \coprod \cdots \coprod Y_N \times Z_N$. We have 
\begin{align*}
    S = S_1 \times_{S_3} S_2 = (X_1 \times Y_1 \times Z_1) \coprod \cdots \coprod (X_N \times Y_N \times Z_N).
\end{align*}
and the maps $\alpha_i \colon S_\nN \to S_i$, $i=1,2$, are induced by a canonical morphism $\gamma \colon S_\nN \to S_1 \times_{S_3} S_2$. Write $\mu = \gamma_\ast T_\nN (a) \in \pP_S.$

We say that the morphisms are \emph{conditionally independent with respect to $\mM_3$} if the following property is satisfied: For any given choice of partitions, as above, it holds that for each index $1 \leq j \leq N$, the probability measures 
\begin{align*}
    \hat{\mu}_1^j := \frac{\mu_1|_{Z_j \times X_j}}{\mu_1(Z_j \times X_j)},\ \hat{\mu}_2^j := \frac{\mu_2|_{Z_j \times Y_j}}{\mu_2(Z_j \times Y_j)}
\end{align*} 
are conditionally independent with respect to the probability measure $\hat{\mu}^j := \frac{\mu |_{X_j \times Y_j \times Z_j}}{\mu(X_j \times Y_j \times Z_j)}$. 

Here, conditional independence (cf.~\cite[Section 1]{swart:1996}) means that for any $z \in Z_j$ and any measurable subsets $U \subseteq X_j, V \subseteq Y_j$ we have
\begin{align} \label{eq:cond ind cont}
    \mathbb{P}_{\hat{\mu}^j}(U \times V \ | \ Z_j = z) = \mathbb{P}_{\hat{\mu}_1^j}(U \ | \ Z_j = z) \cdot \mathbb{P}_{\hat{\mu}_2^j}(V \ | \ Z_j = z).
\end{align}
We are being informal here with the notation used referring to probability density functions (without loss of generality) and assuming that the measures we are dividing by are always nonzero.

\begin{remark}\label{rem:gen}
Definition~\ref{condind} is a generalization of the discrete case. Consider discrete variables $X_j, Y_j, Z_j$. In this scenario, the conditional probabilities can be expressed as follows
    \begin{align*}
        \mathbb{P}_{\hat{\mu}^j}(X_j = u , Y_j = v \ | \ Z_j = z) & = \frac{\mathbb{P}_{\hat{\mu}^j}(X_j = u , Y_j = v, Z_j = z)}{\mathbb{P}_{\hat{\mu}^j}(Z_j = z)} \\
        & = \frac{\mathbb{P}_{\hat{\mu}^j}(X_j = u , Y_j = v, Z_j = z)}{\mu_3(z)} \cdot \mu_3(Z_j),\\
        \mathbb{P}_{\hat{\mu}_1^j}(X_j = u \ | \ Z_j = z) & = \frac{\mathbb{P}_{\hat{\mu}_1^j}(X_j = u, Z_j = z)}{\mathbb{P}_{\hat{\mu}_1^j}(Z_j = z)} = \frac{\mu_1(u,z)}{\mu_3(z)} \cdot \frac{\mu_3(Z_j)}{\mu_1(X_j \times Z_j)} = \frac{\mu_1(u,z)}{\mu_3(z)}, \\
        \mathbb{P}_{\hat{\mu}_2^j}(Y_j = v \ | \ Z_j = z) & = \frac{\mathbb{P}_{\hat{\mu}_2^j}(Y_j = v, Z_j = z)}{\mathbb{P}_{\hat{\mu}_2^j}(Z_j = z)} = \frac{\mu_2(v,z)}{\mu_3(z)} \cdot \frac{\mu_3(Z_j)}{\mu_2(Y_j \times Z_j)} = \frac{\mu_2(u,z)}{\mu_3(z)}.
    \end{align*}
Equality~\eqref{eq:cond ind cont} then becomes
\begin{align*}
  \hat{\mu}^j (u,v,z) = \mathbb{P}_{\hat{\mu}^j}(X_j = u , Y_j = v, Z_j = z) = \frac{\mu_1(u,z)\mu_2(v,z)}{\mu_3(z)\mu_3 (Z_j)} .
\end{align*}
By definition,
\begin{align*}
  T_\nN(a)(\alpha_1^{-1}(u,z) \cap \alpha_2^{-1}(v,z)) = \mu(u , v, z) = \mu(X_j \times Y_j \times Z_j) \cdot \hat{\mu}^j (u,v,z),
\end{align*}
so, using that $\mu_3(Z_j) = \mu_2(Y_j \times Z_j) = \mu_1(X_j \times Z_j) = \mu(X_j \times Y_j \times Z_j)$, we obtain 
\begin{align*}
  T_\nN(a)(\alpha_1^{-1}(u,z) \cap \alpha_2^{-1}(v,z)) = \mu(u , v, z) = \frac{\mu_1(u,z)\mu_2(v,z)}{\mu_3(z)},
\end{align*}
which recovers equation~\eqref{condition for independence}.
\end{remark}

\end{definition}

The following is now a direct consequence of our discussion and work so far.

\begin{theorem} \label{fiber product local fib}
Assume that the state spaces $S_1, S_2$ and $S_3$ are Polish spaces and the morphisms $\mM_1 \to \mM_3$ and $\mM_2 \to \mM_3$ are local fibrations. Then there is a unique probability measure $\mu = T(a)$, giving rise to an MDP $\mM = \mM_1 \times_{\mM_3} \mM_2$ fitting in a commutative diagram~\eqref{fiber product of MDP} such that the morphisms $\mM \to \mM_1$ and $\mM \to \mM_2$ are conditionally independent with respect to $\mM_3$. Moreover, diagram \eqref{fiber product of MDP} is universal among diagrams~\eqref{commutative square of MDP} where the morphisms $\nN \to \mM_i$ form a conditionally independent pair.
\end{theorem}

\begin{proof}
As above, we may assume that $S_3 = Z_1 \coprod \cdots \coprod Z_N$ is a common partition for the fibration structure of the maps $f_i$ with the measures determined by $a \in \Astar$, so that $S_1 = X_1 \times Z_1 \coprod \cdots \coprod X_N \times Z_N$ and $S_2 = Y_1 \times Z_1 \coprod \cdots \coprod Y_N \times Z_N$.

As sets, we then have
\begin{align*}
    S_1 \times_{S_3} S_2 = (X_1 \times Y_1 \times Z_1) \coprod \cdots \coprod (X_N \times Y_N \times Z_N)
\end{align*}
and thus we may reduce to the case $S_3 = Z$ and $S_1 = X \times Z$ and $S_2 = Y \times Z$.

Existence and uniqueness then follows from the existence and uniqueness of $\mu$ with respect to local fibrations $\mM_i \to \mM_3$ \citep{swart:1996,brandenburger:2016} and the normalizing argument of remark~\ref{rem:gen}. Universality can be shown by the continuous version of the argument in the proof of Proposition~\ref{prop:discrete fiber prod thm}.
\end{proof}

\begin{remark}
A Polish space is a separable completely metrizable topological space. The requirement in the preceding theorem that the state spaces be Polish spaces is made for technical reasons and does not affect the substance of our results or their practicability.
\end{remark}

 The constructions in the three previous special cases are applications of Theorem~\ref{fiber product local fib}:
\begin{enumerate}
    \item For the subprocess case, observe that any injection $f_i \colon S_i \to S_3$ is a local fibration by taking $S_i \simeq \left( S_i \times \lbrace \bullet \rbrace \right) \coprod \left( (S_3 \setminus S_i) \times \emptyset \right)$ and $S_3 = S_i \coprod (S_3 \setminus S_i)$. $S_i \times \lbrace \bullet \rbrace \to S_i$ is a fibration with fiber $\lbrace \bullet \rbrace$ and $(S_3 \setminus S_i) \times \emptyset \to S_3 \setminus S_i$ is a fibration with empty fiber.
    
    \item For the finite case, any morphism $f \colon X \to Z$ between finite spaces is a local fibration, because
    $$X = \coprod_{z \in Z} f^{-1}(z) \simeq \coprod_{z \in Z} f^{-1}(z) \times \lbrace z \rbrace$$
    and $f^{-1}(z) \times \lbrace z \rbrace \mapsto z \in Z$ is a fibration with fiber $f^{-1}(z)$.
    
    \item When $S_3$ is finite, any measurable function $f_i \colon S_i \to S_3$ is a local fibration
    $$S_i = \coprod_{s_3 \in S_3} f_i^{-1}(s_3) \simeq \coprod_{s_3 \in S_3} f_i^{-1}(s_3) \times \lbrace s_3 \rbrace$$
    and $f_i^{-1}(s_3) \times \lbrace s_3 \rbrace \mapsto s_3 \in S_3$ is a fibration with fiber $f^{-1}(s_3)$.
\end{enumerate}

\subsection{Pushouts: A Gluing Construction}\label{sec:gluing}

Having discussed pullbacks, we move on to the dual categorical notion of pushout. The pushout models gluing two objects along a third object with morphisms to each. Its universal property is determined by it being minimal in an appropriate sense. Coproducts give a standard example of a pushout. In the category of sets, an example is given by the disjoint union $S_1 \coprod S_2$, which can be viewed as the pushout of the two morphisms $\emptyset \to S_1$ and $\emptyset \to S_2$. 

Intuitively, the pushout is the result of gluing two MDPs $\mM_1$ and $\mM_2$ along a third MDP $\mM_3$ which is expressed as a component of both through morphisms $m_1 \colon \mM_3 \to \mM_1$ and $m_2 \colon \mM_3 \to \mM_2$.

We show that the category $\mathsf{MDP}$ admits pushouts. Unlike the case of pullbacks, we do not encounter measure-theoretic obstructions, as pushouts of sets and pushforwards of measures are compatible (and both covariant) operations. Therefore, $\mathsf{MDP}$ is suitable for expressing compositional structures of MDPs in a universal way.

\begin{theorem}\label{theorem:pushout}
Suppose that we have three MDPs $\mM_1, \mM_2, \mM_3$ together with two morphisms $m_1 = (f_1, g_1) \colon \mM_3 \to \mM_1$ and $m_2 = (f_2, g_2) \colon \mM_3 \to \mM_2$. There exists an MDP $\mM = \mM_1 \cup_{\mM_3} \mM_2$ and morphisms $\mM_1 \to \mM, \mM_2 \to \mM$ fitting in a pushout diagram in $\mathsf{MDP}\textup{:}$
\begin{align*}
    \xymatrix{
    \mM_3 \ar[r]^-{m_1} \ar[d]_-{m_2} & \mM_1 \ar[d] \\
    \mM_2 \ar[r] & \mM
    }
\end{align*}
\end{theorem}

\begin{proof}
We wish to glue $\mM_1$ and $\mM_2$ along their overlap coming from $\mM_3$ to obtain a new MDP, denoted by $\mM \coloneqq \mM_1 \cup_{\mM_3} \mM_2$, such that there exist natural maps $\mM_1 \to \mM$ and $\mM_2 \to \mM$, giving a pushout diagram as above
\begin{align*}
    \xymatrix{
    \mM_3 \ar[r]^-{m_1} \ar[d]_-{m_2} & \mM_1 \ar[d] \\
    \mM_2 \ar[r] & \mM.}
\end{align*}

We propose the following construction (\cref{fig:gluing}).
To define the state space $S$ of $\mM$, we take
\begin{align}
    S = S_1 \cup_{S_3} S_2 = S_1 \coprod S_2 / \sim
\end{align}
where the equivalence relation $\sim$ is generated by identifying $f_1 (s_3) \in S_1$ with $f_2 (s_3) \in S_2$ for all $s_3 \in S_3$. This gives a pushout diagram in the category of sets: 
\begin{align}
    \xymatrix{
    S_3 \ar[r]^{f_1} \ar[d]_-{f_2} & S_1 \ar[d]^-{i_1} \\
    S_2 \ar[r]_-{i_2} & S.
    }
\end{align}
Observe that the state space is a disjoint union of three components
\begin{align} \label{state space of pushout}
    S = (S_1 \setminus f_1(S_3)) \coprod (f_1(S_3) \sim f_2(S_3)) \coprod (S_2 \setminus f_2(S_3)).
\end{align}
Set $S_1^\circ = S_1 \setminus f_1(S_3)$, $S_2^\circ = S_2 \setminus f_2(S_3)$. By abuse of notation, we write $S_3$ to denote the middle component, when this is clear from context.

To specify the action space $\Astar$ and the projection map $\psi \colon \Astar \to S$, by formula \eqref{action space from fibers}, it suffices to define the action spaces $\Astar_s$ for each state $s \in S$. We consider each component separately:
\begin{enumerate}
    \item Over each $S_i^\circ$, for $s \in S_i^\circ$, since $S_i^\circ$ is naturally a subset of $S_i$, we define $\Astar_{s} \coloneqq (\Astar_i)_s$.
    \item Over the overlap $S_3$, for each state $s = f_1(s_3) = f_2(s_3) \in S$, where $s_3 \in S_3$, we let 
    \begin{align} \label{action space on overlap}
    \Astar_{s} = 
    (\Astar_1)_s \coprod (\Astar_2)_{s} / \sim
    \end{align}
    where $\sim$ is generated by identifying $g_1(a_3)$ with $g_2(a_3)$ for all actions $a_3 \in \Astar_3$.
     
\end{enumerate}

\begin{figure}[!t]
  \centering
  \includegraphics[width=.45\linewidth]{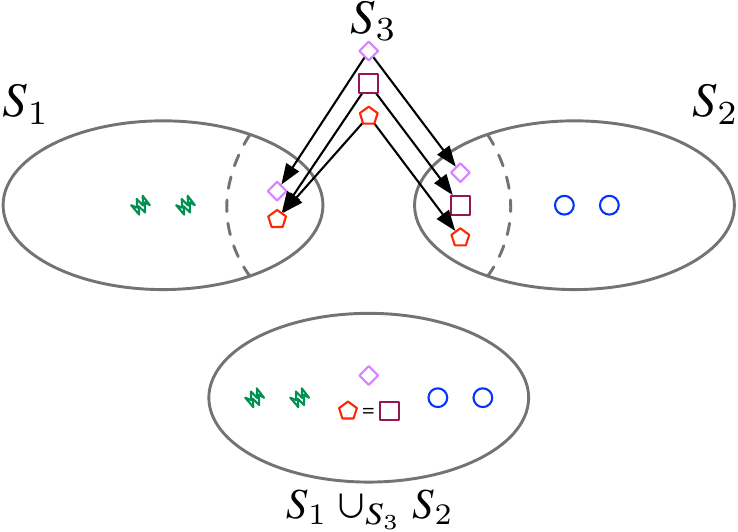}
  \caption{Gluing state spaces $S_i$. Gluing works similarly for state-action spaces $A_i$.}\label{fig:gluing}
\end{figure}

As for state spaces, we have defined $\Astar$ as the following pushout in the category of sets:
\begin{align} 
    \xymatrix{
    \Astar_3 \ar[r]^-{g_1} \ar[d]_-{g_2} & \Astar_1 \ar[d]^-{j_1} \\
    \Astar_2 \ar[r]_{j_2} & \Astar.
    }\label{state-action space of pushout}
\end{align}

By the universal property of pushouts, diagrams~\eqref{state space of pushout} and~\eqref{state-action space of pushout} imply the existence of the canonical morphism $\psi \colon \Astar \to S$.

We move on to the transition probability information, namely the morphism $T \colon \Astar \to \pP_S$. We proceed componentwise specifying $T_s \colon \Astar_s \to \pP_S$ for $s$ in $S_i^\circ$~and~$S_3$:
\begin{enumerate}
    \item Over each $S_i^\circ$, we define $T_s = (T_i)_s$. This makes sense since $\Astar_s = (\Astar_i)_s$ in this case.
    \item Over $S_3$, according to formula \eqref{action space on overlap}, the action space consists of components $(\Astar_i)_s$. 
    
    For $a \in (\Astar_i)_s$ which does not lie in the image of $g_i$, define $T(a)$ to equal $(i_i)_\star T_i(a)$. 
    
    Otherwise, suppose that $a_1 = g_1(a_3) \in \Astar_s $ for some action $a_3 \in \Astar_3$. Write $a_2 = g_2(a_3)$. We observe that
    \begin{align}
        (i_1)_\star T_1(a_1) = (i_2)_\star T_2 (a_2).
    \end{align}
    This follows from the equality $i_1 \circ f_1 = i_2 \circ f_2$ and diagram~\eqref{action-prob compat diagram}, since
    \begin{align*}
        (i_1)_\star T_1(a_1) &= (i_1)_\star T_1(g_1(a_3))\\ 
        &= (i_1)_\star (f_1)_\star T_3(a_3)\\
        &= (i_1 \circ f_1)_\star T_3(a_3), \\
        (i_2)_\star T_2(a_2)
        &= (i_2)_\star T_2(g_2(a_3))\\
        &= (i_2)_\star (f_2)_\star T_3(a_3)\\
        &= (i_2 \circ f_2)_\star T_3(a_3).
    \end{align*}
    We may, thus, define unambiguously
    \begin{align}
        T(a) = (i_1)_\star T_1(a_1) = (i_2)_\star T_2(a_2).
    \end{align}
    This expression is independent of the choice of index $i$, so it respects the equivalence relation $\sim$ on $\Astar$. If the action is in the image of $g_2$, we argue in an identical way. %

\end{enumerate}

Finally, the reward function $R \colon A \to \bR$ is defined to be $R_1$ on the image $j_1(A_3) \subseteq A$ and $R_2$ on the image $j_2(A_3)$. This is well-defined since by definition $R_1 \circ g_1 = R_3 = R_2 \circ g_2$.\\

We now verify that this construction indeed gives a pushout.%

By definition, we have commutative diagrams
\begin{align}
    \xymatrix{
    \Astar_1 \ar[r]^-{j_1} \ar[d]_-{T_1} & \Astar \ar[d]^-{T} \\
    \pP_{S_1} \ar[r]_-{(i_1)_\star} & \pP_S,
    } \quad \xymatrix{
    \Astar_2 \ar[r]^-{j_2} \ar[d]_-{T_2} & \Astar \ar[d]^-{T} \\
    \pP_{S_2} \ar[r]_-{(i_2)_\star} & \pP_S.
    }
\end{align}
Therefore we obtain natural morphisms $\rho_1 = (i_1, j_1) \colon \mM_1 \to \mM$ and $\rho_2 = (i_2, j_2) \colon \mM_2 \to \mM$ fitting in a commutative diagram
\begin{align*}
    \xymatrix{
    \mM_3 \ar[r]^-{m_1} \ar[d]_-{m_2} & \mM_1 \ar[d]^-{\rho_1} \\
    \mM_2 \ar[r]_-{\rho_2} & \mM.
    }
\end{align*}

Now let $\nN$ be a MDP fitting in a commutative diagram
\begin{align*}
    \xymatrix{
    \mM_3 \ar[r]^-{m_1} \ar[d]_-{m_2} & \mM_1 \ar[d]^-{(\alpha_1, \beta_1)} \\
    \mM_2 \ar[r]_-{(\alpha_2, \beta_2)} & \nN.
    }
\end{align*}
We need to check that the diagram is induced by a canonical morphism $m=(f,g) \colon \mM \to \nN$. By the definition of the state and action spaces of $\mM$, they are the pushouts (in the category of sets) of the corresponding spaces of $\mM_i$ along those of $\mM_3$ so there are natural candidates for $f$ and $g$. These fit in a commutative diagram~\eqref{action-space compat diagram}, so it remains to verify that diagram~\eqref{action-prob compat diagram} is commutative, as compatibility of rewards is again clear.

To avoid confusion, we now use the subscripts $\mM$ and $\nN$ to indicate the MDP to which state and action spaces correspond below. 

Since the maps $j_1 \colon \Astar_1 \to \Astar$ and $j_2 \colon \Astar_2 \to \Astar$ are jointly surjective, we only check that the outer square in the diagram
\begin{align*}
    \xymatrix{
    \Astar_1 \ar[r]^-{j_1} \ar[d]_-{T_1} & \Astar_{\mM} \ar[d]^-{T_{\mM}} \ar[r]^-{g} & \Astar_{\nN} \ar[d]^-{T_{\nN}} \\
    \pP_{S_1} \ar[r]_-{(i_1)_\star} & \pP_{S_\mM} \ar[r]_-{f_\star} & \pP_{S_{\nN}}
    }
\end{align*}
commutes and the same is true for the corresponding diagram for $j_2$. But this is the case by construction of $f$ and $g$, since $g \circ j_1 = \beta_1$ and $f \circ i_1 = \alpha_1$ and $(\alpha_1, \beta_1) \colon \mM_1 \to \nN$ is a morphism in $\mathsf{MDP}$.
\end{proof}

The following proposition shows that gluing behaves well with respect to subprocesses.

\begin{proposition}\label{prop:gluing}
Suppose that $\mM_3$ is a subprocess of $\mM_1$ and $\mM_2$. Then $\mM_1$ and $\mM_2$ are subprocesses of $\mM_1 \cup_{\mM_3} \mM_2$.
\end{proposition}

\begin{proof}
We need to show that functions $i_1, i_2$ in diagram~\eqref{state space of pushout} are injective, given that $f_1, f_2$ are injective, and that $j_1, j_2$ in diagram~\eqref{state-action space of pushout} are injective given the injectivity of $g_1, g_2$.
We check that this is true for $i_1$. The argument works for all maps.

Suppose that $i_1 (s_1) = i_1(s_1')$. If $i_1(s_1) \in S_1^\circ \subseteq S$, then we must have $s_1 = s_1' \in S_1^\circ$ since $i_1|_{S_1^\circ}$ is the identity map from $S_1^\circ \subseteq S_1$ to $S_1^\circ \subseteq S$.

If $i_1(s_1) \in f_1(S_3) \subseteq S$, then there exists a unique $s_3 \in S_3$ such that $s_1 = f_1(s_3)$, since $f_1$ is injective. Similarly $s_1' = f_1(s_3')$. Now $i_1(s_1) = i_1(s_1')$ implies that $f_1(s_3) \sim f_1(s_3')$ under the equivalence relation $\sim$ on $S_1 \coprod S_2$, whose quotient is $S$ by definition. Since $f_2$ is also injective, it follows that the equivalence classes of $\sim$ are pairs $\lbrace f_1(t) , f_2(t) \rbrace \subseteq S_1 \coprod S_2$ where $t$ runs through $S_3$. Hence $f_1(s_3) \sim f_1(s_3')$ is only possible if $f_1(s_3) = f_1(s_3')$, which implies that $s_3 = s_3'$.
\end{proof}

We finally state the following corollary, which is a consequence (and generalization) of Theorem~\ref{theorem:pushout} combined with~\cite[\href{https://stacks.math.columbia.edu/tag/002Q}{Lemma 002Q}]{stacks-project} and Proposition~\ref{prop:empty-const}. Namely, a category that has an initial object and admits pushouts, admits finite colimits. These should be thought of as generalizations of pushouts and allow for more complex combinations of MDPs along components.

\begin{corollary}
    The category $\mathsf{MDP}$ admits finite colimits.
\end{corollary}

\section{Compositional Knowledge Representations for Reinforcement Learning}

In this section, we use the compositional theory and its properties to synthesize increasingly complex behaviors. One of the advantages of our framework is the ability to address different forms of compositionality in a uniform and systematic fashion. For example, consider functional (or logical) composition and temporal composition. An instance of the former could be the simultaneous execution of two tasks by an agent, which may or may not interact with each other, such as juggling three balls while riding a unicycle. An instance of the latter could be the sequential completion of two tasks, the first of which might be a prerequisite for the second, such as baking bread before making a sandwich. From the point of view of the theory developed in this paper, both of these compositional behaviors are treated in the same way by the appropriate diagrammatic language (indeed, zig-zag diagrams encapsulate sequential task completion). 

\subsection{Safe Grid Worlds: A Motivating Example}\label{app:grid}

 We consider the case of a grid world~\citep{leike:2017} constructed as a $4\times4$ grid, where an agent attempts to navigate from a starting position to a destination position in the presence of some obstacles (yellow, green and red respectively in \cref{fig:gridworld}).

Definition~\ref{subprocess} is also well-suited to modeling the removal of a subset $\bO \subseteq S$ from the state space of a MDP to obtain a new MDP. 

\begin{figure}[!t]
    \centering
    \includestandalone[width=.6\linewidth]{figures/gridworld}
    \caption{A grid world with a starting state (yellow, $s$), a destination state (green, $\mathbb{D}$), and obstacles (red, $\mathbb{O}$). %
    We can have arbitrary complex worlds of this form; we use the simplest to explain some of the properties we can exploit using categorical RL.}
    \label{fig:gridworld}
\end{figure}

\begin{definition}[Puncturing undesired states and actions]\label{def:punch}
Let $\mM = (S,\Astar, \psi, T, R)$ be an MDP and $\bO \subseteq S$. The MDP $\mM^\circ$ obtained by puncturing $\mM$ along $\bO$ is the MDP $(S^\circ, A^\circ, \psi^\circ, T^\circ, R^\circ)$ where
\begin{enumerate}
    \item $S^\circ = S \setminus \bO$.
    \item Let $B = \lbrace a \in \Astar \ \vert \ T(a)(\bO) > 0 \rbrace$. Then $A^\circ = \Astar \setminus \left( \psi^{-1}(\bO) \cup B \right)$.
    \item $\psi^\circ = \psi|_{A^\circ}$.
    \item $T^\circ = T|_{A^\circ}$.
    \item $R^\circ = R|_{S^\circ}$.
\end{enumerate}
There is a canonical morphism $\mM^\circ \to \mM$, which exhibits $\mM^\circ$ as a subprocess of $\mM$. In fact, $\mM^\circ$ is the canonical maximal subprocess for the subset $S^\circ \subseteq S$. For a morphism of MDPs $m = (f,g) \colon \mM_1 \to \mM_2$, the punctured MDP $\mM_2^\circ$ obtained by puncturing $\mM_2$ along $\mM_1$ is defined as the puncture of $\mM_2$ along the subset $f(S_1) \subseteq S_2$.
\end{definition}

This construction is well-defined since for any action in $A^\circ$, the probability of ending in $\bO$ is zero, as we removed exactly these actions, which forms set $B$. This is where allowing the action spaces to vary along the state space $S$ gives us the flexibility to modify the actions locally to avoid the set $\bO$.

Gluing (Section~\ref{sec:gluing}) also behaves well with respect to puncturing (Proposition~\ref{prop:gluing} and Proposition~\ref{prop:disjoint}).

\begin{proposition}\label{prop:disjoint}
Suppose $\mM_3$ is a subprocess of $\mM_1$ and $\mM_2$ and any action $a_2 \in \Astar_2 \setminus g_2(\Astar_3)$ is not supported on $S_3$, meaning that there is some measurable subset $U_2 \subseteq S_2$ disjoint from $f_2(S_3)$ such that $T(a_2)(U_2) > 0$. Let $\mM_2^\circ$ be the MDP obtained by puncturing $\mM_2$ along $\mM_3$. Then the MDP $(\mM_1 \cup_{\mM_3} \mM_2)^\circ$ obtained by puncturing $\mM_1 \cup_{\mM_3} \mM_2$ along the subprocess $\mM_2^\circ$ is the MDP $\mM_1$.
\end{proposition}
\begin{proof}
The state spaces of $(\mM_1 \cup_{\mM_3} \mM_2)^\circ$ and $\mM_1$ coincide. For the action spaces, by the given condition it follows that puncturing $\mM_1 \cup_{\mM_3} \mM_2$ along $\mM_2$ will remove precisely the actions $A_2 \setminus g_2(A_3) \subseteq A$. But, since $\mM_3$ is a subprocess of $\mM_1$ and $\mM_2$ and $g_1(A_3) = g_2(A_3)$, it follows that $A \setminus (A_2 \setminus g_2(A_3)) = A_1$.
\end{proof}

We illustrate the naturality of the above constructions in the context of puncturing MDPs to enforce a safety condition.
\begin{example}[Static obstacles]
Fix an MDP $\mM = (S, \Astar, \psi, T, R)$ and consider two disjoint subsets $\bO_1, \bO_2 \subseteq S$. We can then construct the MDPs:
\begin{enumerate}
    \item The punctured MDPs $\mM_i^\circ$ along $\bO_i$ for $i=1,2$.
    \item The punctured MDP $\mM_{12}^\circ$ along $\bO_1 \cup \bO_2$.
\end{enumerate}
\end{example}

\begin{proposition}
There exists a commutative diagram%
\begin{align*}
    \xymatrix{\mM^\circ_{12} \ar[r] \ar[d] & \mM^\circ_1 \ar[d] \\
    \mM^\circ_2 \ar[r] & \mM}
\end{align*}
which is simultaneously a pullback and pushout diagram in $\mathsf{MDP}$; that is, $\mM^\circ_{12} = \mM^\circ_1 \times_\mM \mM^\circ_2$ and $\mM = \mM^\circ_1 \cup_{\mM^\circ_{12}} \mM^\circ_2$.
\end{proposition}

\begin{proof}
To see that this is a pullback diagram, observe that the state space of the pullback is the intersection of state spaces of the two punctured MDPs $\mM_i^\circ$, which is the same as the state space of $\mM_{12}^\circ$. For the action spaces, the action space of the pullback consists of the actions in $\mM$ that avoid the two obstacles $\bO_i$. This coincides with the MDP $\mM_{12}$ actions. The transition probabilities coincide tautologically.

The reasoning for the pushout is analogous.
\end{proof}

\begin{example}[Collision avoidance]
Fix an MDP $\mM_1 = (S, \Astar, \psi, T, R)$, which we consider as a model for an agent moving through the state space $S$ with possible actions $A$.
We may work with the product $\mM_2 = \mM \times \mM$ to model the movement of two independent agents.
If, in addition, we would like to make sure that the two agents never collide, we may puncture $\mM \times \mM$ along the diagonal
$$\bO_2 = \Delta = \lbrace (s,s) \ \vert \ s \in S \rbrace \subseteq S \times S$$
to get the MDP $\mM_2^\circ$.
\end{example}

We can use, for example, the above construction to model the movement of $N$ independent agents, ensuring that no two of them collide, by puncturing the product MDP with $N$ factors $\mM_N = \mM \times \ldots \times \mM$ along the big diagonal
\begin{align*}
    \bO_N &= \bigcup_{1\leq i < j \leq N} \Delta_{ij}\\
    &= \bigcup_{1\leq i < j \leq N} \lbrace (s_1, \ldots, s_N) \ \vert \ s_i = s_j \rbrace \subseteq S^N
\end{align*}
to define the MDP $\mM_N^\circ$.

\subsection{Zig-zag Diagrams: A Language Equipped with Compositional Verification}\label{sub:zig-zag}
\begin{figure}[!t]
    \centering
    \includegraphics[width=.8\textwidth]{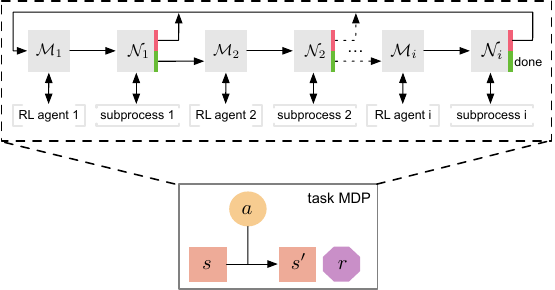}
    \caption{Main task decomposed into sub-tasks ($M_1, M_2, \cdots, M_i$) and subprocesses that check for correct task completion ($N_1, N_2, \cdots, N_i$): the categorical formalism introduces compositional guarantees for the chaining of tasks.}
    \label{fig:decomp}
\end{figure}

For designing compositional tasks, we desire to operationalize using the categorical semantics of RL, that involve accomplishing tasks sequentially. The zig-zag diagrams we consider below are not merely an intuitive diagrammatic representation of sequential task completion, but rather they invoke directly the results of this work. Therefore, they represent a form of \emph{compositional verification} for operational goals and stopping conditions (Figure~\ref{fig:decomp}).

In a general setting, we consider the setup given by, what we term, a \emph{zig-zag} diagram of MDPs
    \begin{align} \label{zig-zag}
        \xymatrix@R-=0.3cm@C-=0cm{
        & \nN_0 \ar[dl] \ar[dr] & & \nN_1 \ar[dl] \ar[dr] & & \ldots & & \nN_{n-1} \ar[dr] \ar[dl] \\
        \mM_0 & & \mM_1 & & \mM_2 & \ldots & \mM_{n-1} & & \mM_n
        }
    \end{align}
where for each $i=0, \ldots, n-1$, $\nN_i$ is a subprocess of $\mM_i$ (Definition~\ref{subprocess}).

The composite MDP associated to the above diagram is the MDP $\cC_n$ defined by the inductive rule
\begin{align*}
    \cC_0  &\coloneqq \mM_0,\\
    \cC_1  &\coloneqq \cC_0 \cup_{\nN_0} \mM_1,\\
    & \qquad \vdots\\
    \cC_n  &\coloneqq \cC_{n-1} \cup_{\nN_{n-1}} \mM_n.
\end{align*}

The intuitive interpretation of the above zig-zag diagram is that it blackboxes compositional task completion. In particular, each subprocess $\nN_i \to \mM_i$ models the completion of a task in the sense that the goal of an agent is to eventually find themself at a state of $\nN_i$. 
Once the $i$-th goal is accomplished inside the environment given by $\mM_i$, we allow for the possibility of a changing environment and more options for states and actions in order to achieve the next goal modeled by the subprocess $\nN_{i+1} \to \mM_{i+1}$.

The composite MDP $\cC_n$ is a single environment capturing all the tasks at the same time (and is in fact the colimit of diagram~\eqref{zig-zag} in the category $\mathsf{MDP}$).
\begin{itemize}[leftmargin=2.5em,align=left]
    \item[$\left<?\right>$]  Suppose an agent has learned an optimal policy for each MDP $\mM_i$ given the reward function $R_i$ for achieving the $i$-th goal for each $i=0, \dots,n$. Under what conditions do these optimal policies determine optimality for the joint reward on the composite MDP $\cC_n$? 
\end{itemize}

A scenario in which this is true is when the zig-zag diagram is forward-moving, meaning that $\nN_i$ is a full subprocess of $\mM_i$, and the optimal value function $v_\star (s)$ for any state $s$ in the state space of a component $\mM_i$, considered as a state of $\cC_n$, is \emph{monotonic} for subsequent subprocesses $\mM_{i+1}, \ldots, \mM_n$. For simplicity, we take everything to be discrete. Monotonicity here means that the expressions
\begin{align*}
    \sum_{s' \in S_i} T(a)(s') (R_i(a) + \gamma \cdot v_\star^{\cC_n} (s') ) \\
    \sum_{s' \in S_i} T(a)(s') (R_i(a) + \gamma \cdot v_\star^{\cC_{[i,n]}} (s') )
\end{align*}
are maximized by the same action $a \in (A_i)_s$, where we have fixed a discount factor $\gamma$. Here $\cC_{[i,n]}$ denotes the composite MDP of the truncated zig-zag diagram
\begin{align*}
        \xymatrix@R-=0.3cm@C-=0cm{
        & \nN_i \ar[dl] \ar[dr] & & \nN_{i+1} \ar[dl] \ar[dr] & & \ldots & & \nN_{n-1} \ar[dr] \ar[dl] \\
        \mM_i & & \mM_{i+1} & & \mM_{i+2} & \ldots & \mM_{n-1} & & \mM_n.
        }
\end{align*}
A zig-zag diagram can always be made forward-moving by removing the actions of $\nN_i$ that can potentially move the agent off $\nN_i$ back into $\mM_i$. This is the operation of puncturing $\mM_i$ along the complement of $\nN_i$ and intersecting the result with $\nN_i$.

\begin{theorem} \label{Theorem 3}
Suppose that a zig-zag diagram is forward-moving and the optimal value function of $\cC_n$ is monotonic as above. Then, following the optimal policy $\pi_i$ on each component $\mM_i$ gives an optimal policy on the composite MDP $\cC_n$.
\end{theorem}

\begin{proof}
First, consider the Bellman equation for all $s \in S$, $a\in A$, $R$ the reward function, and $\gamma$ the discount factor we have fixed~\cite[chapter 4]{sutton:2018}
\begin{equation*}
v_\star (s)=\max_{a \in A_s} \sum_{s' \in S} T(a)(s') \left(R(a)+\gamma v_\star \left(s^{\prime}\right)\right).
\end{equation*}

We argue by reverse induction. Suppose that the claim is true for the composite $\cC_{[i+1,n]}$.

We then show that it is true for $\cC_{[i,n]}$. Since the zig-zag diagram is forward-moving, the optimal values of $\cC_{[i,n]}$ and $\cC_{[i+1,n]}$ coincide on the state space of the latter, as successor states in $\cC_{[i+1,n]}$ are states in $\cC_{[i+1,n]}$. For the same reason, the optimal values of states of $\mM_i$ are the same for the composite MDPs $\cC_n$ and $\cC_{[i,n]}$. The monotonicity condition now implies that following policy $\pi_i$ on $\mM_i$ and the optimal policy on $\cC_{[i+1,n]}$ gives an optimal policy on $\cC_{[i,n]}$.

For the base case, observe that $\mM_n = \cC_{[n,n]}$ which has maximal policy $\pi_n$.
\end{proof}

\begin{example}[Visiting regions sequentially]\label{ex:seq}
Consider a point mass robot sequentially visiting three points, $R_1$, $R_2$, $R_3$, in a grid world, while always avoiding obstacles.
\end{example}

We can model this problem compositionally by the following zig-zag diagram, where we consider punctured MDPs (Proposition~\ref{def:punch})
\begin{align*}
        \xymatrix@R-=0.3cm@C-=0.3cm{
          & & R_1 & & R_2 & & R_3 & & \\
          & & \mathsf{pt} \ar[dl] \ar[dr] & & \mathsf{pt}  \ar[dl] \ar[dr]  &  & \mathsf{pt}  \ar[dr] \ar[dl] & &  \\
        & \mM_1^\circ & & \mM_2^\circ & & \mM_3^\circ &  & \mM_4^\circ.
        }
\end{align*}

At any given position in the grid the five options for the point mass robot are forward, backwards, left, right, and stay in the same position. Here each $\mM_i^\circ$ denotes the MDP in which all the obstacles have been punctured and the actions forward, backwards, left, right have been removed at the point $R_i$. Each intermediate subprocess $\mathsf{pt} \to \mM_i$ maps the stationary point to the point region $R_i$. This requires us moving from MDP to MDP to puncture the actions that lead to moving away from the next subsequent sequence we want.

The problem is inductively defined by the composite MDP: \[\cC_\textrm{robot} = \mM_1^\circ \cup_{\mathsf{pt}} \mM_2^\circ \cup_{\mathsf{pt}} \mM_3^\circ \cup_{\mathsf{pt}} \mM_4^\circ.\]

Observe that this zig-zag diagram is forward-moving and the optimal value function of $\cC_\textrm{robot}$ is monotonic. Thus, the optimal policy on $\cC_\textrm{robot}$ is given by the optimal policy of each component $\mM_i^\circ$ (Theorem~\ref{Theorem 3}).
\smallskip

The more categorically minded reader will observe that zig-zag diagrams bear a close relationship with spans and cospans~\citep{cicala:2016}. %

In practice, the language of zig-zag diagrams in RL enables a structured, abstract, and rigorous understanding of complex sequential tasks via the following properties.
\begin{itemize}
\item \textbf{Semantics:} 
Diagrammatic languages map syntactic constructs to mathematical objects, explaining what each part of a system means~\citep{diskin:2012}. Zig-zag diagrams represent the relationships between MDPs and subprocesses in a sequential task and encode the semantics of how these processes interact.
\item \textbf{Compositionality:} 
The meaning of the constituent parts determines the meaning of a complex expression, ensuring modularity~\citep{coecke:2023}. Similarly, the zig-zag pattern shows how complex processes comprise  subprocesses and individual MDPs.
\item \textbf{Abstraction:} 
Diagrammatic languages abstract away many implementation details and focus on the meaning or behavior of constructs. Zig-zag diagrams abstract away the specific workings of each MDP and subprocess, focusing instead on their high-level relationships. Similar diagrammatic languages from applied category theory have used abstraction to make progress in designing complex systems~\citep{zardini:2021,abbott:2024,bakirtzis:2021,breiner:2019a,schultz:2016,spivak:2016,bonchi:2019,gavranovic:2024,hedges:2024}.
\item \textbf{Formal system:} The relationships represented in a zig-zag diagram are subject to compositionality conditions, providing a formal view to manipulating the composite system often used in, for example, model-based engineering~\citep{diskin:2019}.
\end{itemize}
\subsection{State-action Symmetry}\label{sec:sym}

So far we have discussed approaches to helping designers of RL systems organize into functional subsystems. However, categorical semantics can provide benefits beyond that. By exploiting structure, we can make learning easier for the agents; for example,
one of the benefits of working with algebras in some category is that we can use gadgets from algebra to exploit the geometry of an RL problem. Using algebraic gadgets becomes helpful when considering efficient approaches to \emph{symmetric} RL problems, such as homomorphic networks~\citep{pol:2020}. We will investigate how homomorphic networks abstract within our categorical theory and improve generalization. 

MDP homomorphisms in our theory can be viewed as morphisms between MDPs, which preserve composition and \emph{forget} unrelated structure. In the concept of symmetry, we will rely on the construction of a \emph{quotient MDP}. The symmetric structure of the RL problems we consider in this section alters the state-action space such that it is more economical to use the geometry of the problem. The construction removes the state-action pairs that can be related through symmetry. These symmetries show up in several RL problems and are common to mechanical systems, where we can think of  symmetry as moving up compared to down or left compared to the right. In the larger context of designing RL systems we can think of the below operations as engineering within an \emph{abstraction}---related to the ``forget'' operation. In some cases, we would instead need to add structure, which would be the same as applying some sort of \emph{refinement} operation within our framework.

The goal of discussing symmetries is, therefore, twofold. First, we study symmetries to show the generality of our categorical formalism. Second, we study symmetries to show how design operations, as they occur in the engineering of RL systems, such as abstraction, reflect precisely within the categorical semantics of RL. 

Denote by $\Aut(\mM)$ the set of isomorphisms of an object of $\mathsf{MDP}$; a group under composition.
A group action of a group $G$ on an MDP $\mM$ is a group homomorphism $\rho \colon G \to \Aut(\mM)$. 
Concretely this means that for every group element $g \in G$ there is an isomorphism $\rho_g = (\alpha_g, \beta_g) \colon \mM \to \mM$ satisfying the composition identity $\rho_{gh} = \rho_g \circ \rho_h$. Intuitively a group action gives a set of symmetries for an MDP $\mM$. We would like to perform RL keeping this mind and obtain policies that are invariant under the given domain symmetries. In order to do this efficiently, we need access to a quotient MDP $\mM / G$.

\begin{figure}[!t]
    \centering
  \includegraphics[width=.4\textwidth]{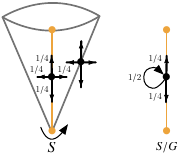}
  \caption{A conical state space collapses to a line in the quotient by axial rotation. The actions of going up and down stay the same, while left and right merge into one.}\label{fig:symmetry}
\end{figure}
For a group $G$, we have a natural quotient MDP $\mM / G$ constructed as follows (for example see  \cref{fig:symmetry}):

\begin{enumerate}
    \item Let $\widehat{\mM} \coloneqq \mM \times G$ be the MDP with state space $\widehat{S} = S \times G$ and action space $\widehat{A} = A \times G$. We have $\widehat{\psi} = \psi \times \id_G$ and the transition probabilities are given by the map
    $$\widehat{T}(a,g) =(\id_S \times \iota_g)_\star T(a)$$
    where $\iota_g \colon S \to G$ denotes the constant map with value $g \in G$.
    \item There are natural group action and projection morphisms $\rho, pr_1 \colon \widehat{\mM} \to \mM$ defined as:
    \begin{itemize}
        \item For $\rho$, the map on state and action spaces is given by the group actions $\alpha \coloneqq S \times G \to S$ and $\beta \coloneqq A \times G \to A$. This gives a morphism of MDPs since for any $(a,g) \in A \times G$ mapping to $(s, g) \in S \times G$ under $\widehat{\psi}$ we have
        \begin{align*}
            \alpha_\star \widehat{T}(a,g) = \alpha_\star (\id_S \times \iota_g)_\star T(a) = (\alpha_g)_\star T(a) = T(\beta_g(a)) = T(\beta(a,g))
        \end{align*}
        where we used that by definition $\alpha \circ (\id_S \times \iota_g) = \alpha_g$.
        \item For $pr_1$, the same argument works for the first projection maps $S\times G \to S$ and $A \times G \to A$.
    \end{itemize}
    \item Define $\mM / G$ as fitting in the pushout diagram
    \begin{align*}
        \xymatrix{
        \mM \times G \ar[r]^-{pr_1} \ar[d]_-{\rho} & \mM \ar[d]^-q \\
        \mM \ar[r]_-q & \mM / G.
        }
    \end{align*}
    Thus $\mM / G \coloneqq \mM \cup_{\mM \times G} \mM$ and $q \colon \mM \to \mM / G$ is the canonical quotient morphism. \end{enumerate}

The following proposition confirms that the quotient $\mM / G$ satisfies the desired universal property.

\begin{proposition}
$\mM/G$ is the quotient of the MDP $\mM$ by the action of $G$ in the sense that for any MDP $\nN$ and a $G$-invariant morphism $m \colon \mM \to \nN$, there is a unique factorization
$\mM \xrightarrow{q} \mM / G \to \nN$.
\end{proposition}
\begin{proof}
This follows immediately from the definition of $\mM / G$. A morphism $m \colon \mM \to \nN$ is $G$-invariant if the maps between their state and action spaces are $G$-invariant and this is equivalent to the condition that
$$m \circ pr_1 = m \circ \rho \colon \mM \times G \to \nN. $$
The conclusion then follows by the universal property of the pushout $\mM / G$.
\end{proof}

\subsection{A Design for Compositional Task Completion: Putting It All Together}\label{sec:design}

\begin{figure}[!t]
    \centering
    \includegraphics[clip, trim=9cm 5cm 9cm 2.5cm, width=.45\linewidth]{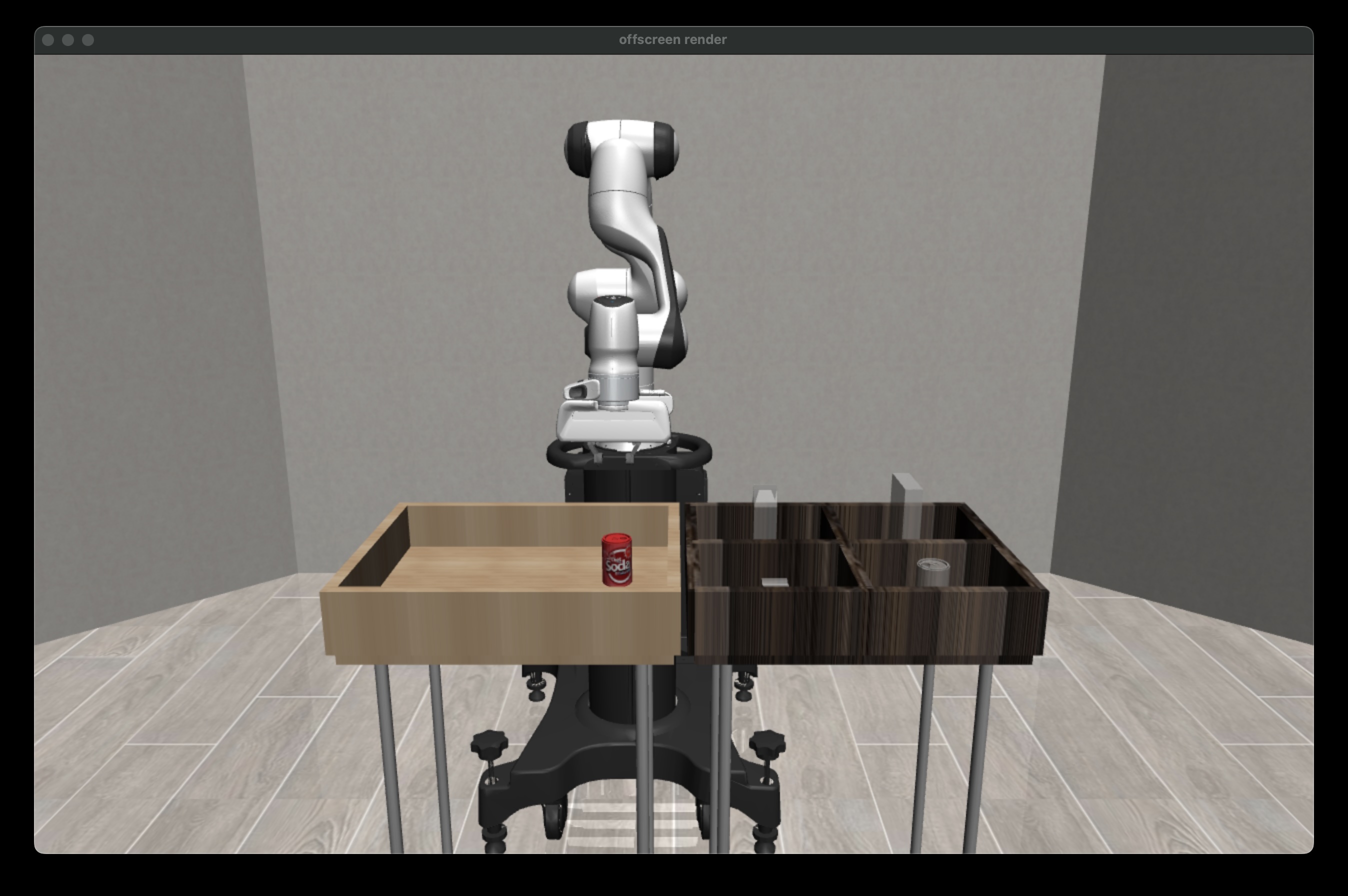}
    \caption{A fetch-and-place robot transfers objects. In dynamic environments, the robot should also be able to handle moving objects within a defined area.}
    \label{fig:fetch}
\end{figure}

Compositional RL problems are \emph{sequential} in nature. An action follows another, given some rules of engagement.%
 Those rules include how actions modify given the presence of another agent or how the actions of agents ought to intertwine in one sequential task description. 
We adapt a fetch-and-place robot problem (\cref{fig:fetch}), where the learning algorithm controls actuation and rewards are assigned at task completion~\citep{sutton:2018}.

\begin{example}[Fetch-and-place robot] \label{robot example}
Suppose that a robotic arm wants to fetch a moving object from inside a box and then place it on a shelf outside the box.

This is captured by the diagram of MDPs
\begin{align*}
    \xymatrix@R-=0.3cm@C-=0.3cm{
    & \mathsf{Fetch} \ar[dl] \ar[dr] & &  \mathsf{Place} \ar[dl] \ar[dr] \\
    \mathsf{Box} & & \mathsf{Move} & & \mathsf{Place}
    }
\end{align*}

We give a simple intuitive description of each:
\begin{enumerate}
    \item [$\mathsf{Box}$] The state space is $B \times B$ where each factor records the position of the robot arm tip and the moving object and the action space is $A \times A$. 
    \item [$\mathsf{Fetch}$] The object has been fetched and, thus, the position and actions of the arm and the object coincide. The state and action spaces are given by the diagonals in $B \times B$ and $A \times A$. More generally, $\mathsf{Fetch}$ is the maximal subprocess associated with the diagonal as a subset of $B \times B$. Observe that we have made our state and action spaces smaller by recording data of half the dimension.
    \item [$\mathsf{Move}$] Since the arm needs to move the object outside of the box, we need to enlarge the state space. Thus, 
    $\mathsf{Move} = \mathsf{Fetch} \cup_{\mathsf{Overlap}} \mathsf{Outside}$
    where $\mathsf{Overlap}$ is a common region of the box and the outside environment. The actions are defined to allow the arm and object to move within the whole environment.
    \item [$\mathsf{Place}$] This is a full subprocess of $\mathsf{Outside} \to \mathsf{Move}$. If the ending position is a point in $\mathsf{Outside}$, then we may take $\mathsf{Place} = \mathsf{pt}$ and a subprocess $\mathsf{pt} \to \mathsf{Outside}$.
\end{enumerate}

The composite MDP in this setup can be expressed as 
\begin{align*}
    \cC &= (\mathsf{Box} \cup_{\mathsf{Fetch}} \mathsf{Move}) \cup_{\mathsf{Place}} \mathsf{Place}\\
    &= \mathsf{Box} \cup_{\mathsf{Fetch}} \mathsf{Move}\\
    &= \mathsf{Box} \cup_{\mathsf{Fetch}} (\mathsf{Fetch} \cup_{\mathsf{Overlap}} \mathsf{Outside}).
\end{align*}
If the object is stationary, we can make the diagram forward-moving by deleting the actions of $\mathsf{Fetch}$ inside $\mathsf{Box}$ which separate the arm and the object. Then, $\mathsf{Fetch}$ is full in $\mathsf{Box}$, but it is no longer full in composite $\cC$, allowing for continuation of movement in order to complete the final task $\mathsf{Place}$ which remains full in $\cC$. In that case, we can apply Theorem~\ref{Theorem 3} to compute the policies componentwise, meaning that we prove that learning-by-parts is the same as learning on the whole given some conditions for the composition of MDPs.
\end{example}

\section{Related Work}

Category theory has found applications across mathematics and computer science, serving as a unifying mathematical language that captures the structure and relationships of systems~\citep{pierce:1991}. In this paper, we give a uniform treatment and provide universal interfaces to abstractions of MDPs and their homomorphisms~\citep{li:2006,ravindran:2003,ravindran:2004}, allowing, among others, a rigorous universal formalization of subtasks and embeddings of MDPs. These generalizations and unifications are particularly compelling for making progress in agent generalization via defining the abstract class of sequential decision-making and letting the agent learn the specifics of the problem~\citep{konidaris:2019} or being explicit in recovering and forgetting information when traversing the RL abstraction hierarchy~\citep{abel:2018}. Our work is supported by previous findings on the complexity and efficiency of hierarchical RL in the context of MDP interactions~\citep{zheng:2020}.  %

Hierarchical decomposition has been a prominent theme in RL, often implemented through hierarchical MDPs~\citep{parr:1997,dietterich:2000,ravindran:2013,nachum:2018}. Unlike previous work focusing on the decomposition of tasks, the categorical approach universally describes the structure of MDPs and their relationships to understand compositionality, scalability, and interoperability in complex decision-making systems. The immediate consequence of the zig-zag diagram is the formal proof of the minimal conditions for reaching a goal from one MDP to another. Finding the right initiation set has been effectively studied for particular classes of problems~\citep{bagaria:2023}, but the categorical formalism results in full generality---we prove the specific rules of engagement that must be satisfied to verify the compositionality of any framework, working in tandem rather than in opposition. \citet{neary:2022} compose MDPs into subtask policies and meta-policies for the purpose of verification, in fact the composition is intuitive and clever in that largely experimental work but the \emph{why} it works fits neatly in the zig-zag diagrams we present in the current paper. 

Building on these principles, compositionality in RL has typically emphasized temporal and state abstractions, executing complex behaviors, and improving learning efficiency through mechanisms such as chaining known skills~\citep{tasse:2020,tasse:2022,niekerk:2019,jothimurugan:2021,ivanov:2021}. However, our categorical formalism shifts the focus toward the functional composition in RL through universal properties. We use the categorical formalism to decompose tasks into behavioral \emph{functions}. Unlike previous work, this functional approach provides a more granular, mathematical structure for task decomposition. Functional composition aligns with recent efforts in solving robotics tasks~\citep{devin:2017} and discovering policy decomposition into modular neural architectures~\citep{mendez:2022}, offering a rigorous compositional framework equipped with a corresponding diagrammatic language of zig-zag diagrams. The relationship between the categorical formalism and the ideas of logical composition is one of coexistence rather than comparison, in the sense that both can make use of the other to verify increasingly complex classes of RL systems. 

Through monads and functors, categorical semantics have been used to describe stochastic processes and model uncertainty within probabilistic computations~\citep{giry:1981,watanabe:2023}. Other work has attempted to apply category theory to model MDPs, embedding the dynamics of decision processes within a categorical framework~\citep{fritz:2023,baez:2016}. While these studies create a bridge between abstract mathematics and probabilistic modeling, our work applies categorical principles to functional decomposition in RL. We extend the categorical perspective beyond mere probabilistic representation, using it to systematically structure complex tasks, compose behavior and  produce functional decision-making algorithms for RL. Other adjacent work relates to using a category-theoretic lens in control, dynamical systems, and robotics~\citep{hanks:2024,ames:2025b,ames:2025a,ames:2006, culbertson:2019,lerman:2018,lerman:2020}.

\section{Conclusion}

This work demonstrates how a compositional theory can be used to 
systematically model complex behaviors, integrating both functional and 
temporal composition within a unified framework. By using a 
consistent diagrammatic approach, we show that seemingly distinct 
behaviors, such as independent task execution (for example, juggling while 
riding a unicycle) and sequential task completion (for example, baking bread 
before making a sandwich), can be treated equivalently. This provides a 
more rigorous and flexible method for analyzing compositional dynamics 
in diverse scenarios.

In general, structural conditions are essential for RL generalization~\citep{du:2021}, safety~\citep{alshiekh:2018}, and sample efficiency~\citep{sun:2019}. Making explicit the relational aspects of these structural conditions gives rise to concrete mappings between formalisms~\citep{bakirtzis:2022a}, a currently open problem in  learning~\citep{lukcuck:2019}. Our theoretical results give meaning to the robustness of RL's compositionality feature for structural conditions. In particular, we prove that the composition of MDPs is a pullback and has a well-defined measure, there exists a gluing operation of MDPs, and that as long as MDP composites are forward-moving, then the learning-by-parts corresponds to learning the optimal policy on the whole. Following the distinction~\citet{pattee:1978} makes between universal laws and local rules, this work seeks to make progress in identifying the fundamental \emph{laws} of reinforcement learning---which are inexorable, incorporeal, and universal---as opposed to the more arbitrary, structure-dependent, and local \emph{rules}. %

\acks{U.T. would like to acknowledge his support from the ARO W911NF-20-1-0140 and AFOSR FA9550-22-1-0403 grants.}

\appendix
\section{Preliminaries}\label{app:cat}

Here we give a brief account of definitions of some of the mathematical structures we use in order to study the universal properties of the compositionality features in RL. 

\subsection{Some Categorical Notions} Consult~\citet{lawvere:2009,leinster:2014}, or~\citet{mac:1998} for an in-depth treatment of category theory.

\begin{definition}[Category]
A (small) \emph{category} $\mathsf{C}$ consists 
of a collection of a set of objects $\mathrm{Ob}$ and for any two $X, Y \in \mathrm{Ob}$ a set of arrows $f\colon X\to Y$, along with a composition rule 
\begin{displaymath}
(f\colon X\to Y,\ g\colon Y\to Z)\mapsto g\circ f\colon X\to Z
\end{displaymath}
and an identity arrow $\id_X\colon X\to X$ for all objects, subject to associativity, whenever compositions make sense, and unity conditions: $(f\circ g)\circ h=f\circ (g\circ h)$ and $f \circ \id_X=f=\id_Y\circ f$.
\end{definition}

This definition encompasses a vast variety of structures 
in mathematics and other sciences: to name a couple, $\mathsf{Set}$ is the category 
of sets and functions between them, whereas $\mathsf{Lin}_k$ denotes the category 
of $k$-linear real vector spaces and $k$-linear maps between them, where $k$ is a given field.

\begin{definition}[Commutative diagrams]
A standard diagrammatic way to express composites is by concatenation of arrows: $X\xrightarrow{f}Y\xrightarrow{g}Z$. One can also express equations between morphisms via commutative triangles of the form
\begin{displaymath}
\begin{tikzcd}
X\ar[r,"f"]\ar[dr,"h"'] & X\ar[d,"g"]\\
& Y
\end{tikzcd}\quad\textrm{ which stands for }g\circ f=h;
\end{displaymath}

or commutative squares of the form

\begin{displaymath}
\begin{tikzcd}
X\ar[r,"f"]\ar[d,"g"'] & Z\ar[d,"g'"]\\
Y\ar[r,"f'"]& W
\end{tikzcd}\quad\textrm{ which stands for }g'\circ f= f' \circ g.
\end{displaymath}

More complicated diagrammatic expressions account for analogous relationships between morphisms.
\end{definition}

\begin{definition}[Isomorphism]
A morphism $f\colon X\to Y$ is called \emph{invertible} or an \emph{isomorphism} when there exists a morphism $g\colon Y\to X$ such that $f\circ g= \id_Y$ and $g\circ f=\id_X$.
\end{definition}

\begin{definition}[Functor]
A \emph{functor} $F\colon\mathsf{C}\to\mathsf{D}$ between two categories consists of a function between objects and a function between morphisms, mapping an object $X$ of $\mathsf{C}$ to an object $FX$ of $\mathsf{D}$ and a morphism $f \colon X \to Y$ to $Ff\colon FX\to FY$, such that it preserves composition and identities: $F(f\circ g)=Ff\circ Fg$ and $F(\id_X)=\id_{FX}$.
\end{definition}

A functor can be informally thought of as a structure-preserving map between domains of discourse.
Interestingly, categories and functors form a category on their own,
denoted $\mathsf{Cat}$, in the sense that functors can be composed and the rest of the axioms hold.

The following definition is more technical and is provided for completeness.

\begin{definition}[Cartesian category]
A category is called cartesian closed if it has all finite products and exponentials.
\end{definition}

The category of sets $\mathsf{Set}$ is cartesian closed where the product is the common product
\begin{equation*}
A \times B=\{(a, b) \mid a \in A, b \in B\}.
\end{equation*}

We can think of this operation as, for example, organizing data on a table and the projections are then giving us the particular column and row respectively.

Pullbacks are a generalization of the notion of a product.

\begin{definition}[Pullback]
A category is said to have a pullback for the pair of morphisms $f\colon A \to C$ and $g\colon B \to C$ when there exists an object $W$ together with morphisms $a\colon W \to A$ and  $b\colon W \to B$ such that the square

\begin{displaymath}
\begin{tikzcd}
W\ar[r,"b"]\ar[d,"a"'] & B\ar[d,"g"]\\
A\ar[r,"f"]& C
\end{tikzcd}
\end{displaymath}
commutes, that is $f \circ a = g \circ b$, and which is universal in the following way: for any object $W_o$ with morphisms $a_o \colon W_o \to A$ and $b_o \colon W_o \to B$ such that $f \circ a_o = g \circ b_o$, there exist a unique morphism $w\colon W_o \to W$ such that $a \circ w = a_o$ and $b \circ w = b_o$.

$W$ is then the limit of the diagram 
\begin{displaymath}
\begin{tikzcd}
 & B\ar[d,"g"]\\
A\ar[r,"f"]& C
\end{tikzcd}
\end{displaymath}
and we write $W = A \times_C B$ for the pullback. We also say that $W$ is the pullback of $A$ along the map $g \colon B \to C$ and also the pullback of $B$ along the map $f \colon A \to C$. $W$ is also the fiber product of the pair of morphisms $f \colon A \to C$ and $g \colon B \to C$.
\end{definition}

In analogy with the product of two sets $A \times B$ with the two projections $A \times B \to A$ and $A \times B \to B$, the morphisms $a$, $b$ are often thought of as projection maps.

\begin{definition}[Pushout]
    A pushout for morphisms $f \colon C \to A$ and $g \colon C \to B$ is an object $W$ together with morphisms $a\colon A \to W$ and  $b\colon B \to W$ such that the square

\begin{displaymath}
\begin{tikzcd}
C\ar[r,"g"]\ar[d,"f"'] & B\ar[d,"b"]\\
A\ar[r,"a"]& W
\end{tikzcd}
\end{displaymath}
commutes, that is $a \circ f = b \circ g$, and which is universal in the following way: for any object $W_o$ with morphisms $a_o \colon A \to W_o$ and $b_o \colon B \to W_o$ such that $ a_o \circ f =  b_o \circ g$, there exist a unique morphism $w\colon W \to W_o $ such that $w \circ a = a_o$ and $w \circ b = b_o$.

$W$ is then the colimit of the diagram 
\begin{displaymath}
\begin{tikzcd}
C\ar[r,"g"]\ar[d,"f"'] & B\\
A& 
\end{tikzcd}
\end{displaymath}
and we write $W = A \cup_C B$ for the pushout.
    
\end{definition}

We see that a pushout is the dual version of a pullback. The pushout in the category $\mathsf{Set}$ of two morphisms $\emptyset \to A$ and $\emptyset \to B$ is the standard disjoint union $A  \coprod B$, labelling the composed set with which elements come from set $A$ and which elements come from set $B$. This is also the coproduct in the category of sets. For another example, when we have a non-empty intersection $A \cap B \subseteq A$ and $A \cap B \subseteq B$ and $a\colon A \cap B \to A$ and $b\colon A \cap B \to B$ are the inclusions, the union $A \cup B$ is naturally isomorphic with the pushout of $a$ and $b$. More generally, the morphisms $a$, $b$ can be thought of as inclusions in keeping with these examples.\\

\subsection{Pushforwards in Measure Theory} For a treatment of measure theory consult~\citet{billingsley:1986}.

\begin{definition}[Pushforward measure]
Given measurable spaces $(\Omega, \Sigma)$ and $(X, T)$, a measure $\mu$ on $(\Omega, \Sigma)$ and a measurable function $\psi \colon (\Omega, \Sigma) \to (X, T)$, we write $\psi_\star \mu$ for the pushforward measure obtained from $\mu$ by applying $\psi$: 
\[\psi_\star \mu (A) \coloneqq \mu\left(\psi^{-1} (A)\right) \text{ for all } A \in T.\]
\end{definition}

For example, for a deterministic function with random inputs, the pushforward measure gives us an explicit description of the possible distribution of outcomes.\\

\subsection{Groups} For an in-depth treatment of algebra consult~\citet{aluffi:2021}.

\begin{definition}[Group]
A set $G$ endowed with a binary operation $\bullet$, $(G, \bullet)$, is a group if the following conditions hold.
\begin{enumerate}
    \item The operation $\bullet$ is associative; that is, $(\mathrm{for\; all}\; g, h, k \in G)\colon (g \bullet h) \bullet k = g \bullet (h \bullet k)$.
    \item There exists an identity element $e_G$ for $\bullet$; that is, $$(\mathrm{there\; exists\;} e_G \in G)(\mathrm{for\; all\;} g \in G)\colon g \bullet e_G = e_G \bullet g.$$
    \item Every element in $G$ is invertible with respect to $\bullet$; that is, $$(\mathrm{for\; all\;} g \in G)(\mathrm{there\; exists\;} h \in G)\colon g \bullet h = h \bullet g = e_G.$$
 \end{enumerate}
\end{definition}

Consider as an example the set of integers $\mathbb{Z}$:
\begin{enumerate}
    \item Addition in $\mathbb{Z}$ is an associative operation.
    \item The identity element is the integer $0$.
    \item The inverse map sends an integer $n \in \mathbb{Z}$ to another integer $-n$.
\end{enumerate}

\bibliography{manuscript}

\end{document}